\documentclass[10pt]{article}
\usepackage[preprint]{jmlr2e}

\usepackage{amsmath,amsfonts,bm}

\def\eqref#1{equation~\ref{#1}}

\def\1{\bm{1}}

\DeclareMathAlphabet{\mathsfit}{\encodingdefault}{\sfdefault}{m}{sl}
\SetMathAlphabet{\mathsfit}{bold}{\encodingdefault}{\sfdefault}{bx}{n}

\usepackage{booktabs}

\usepackage{sidecap} 
\usepackage{url}
\usepackage{algorithm}
\usepackage{graphicx}
\usepackage{subcaption}
\usepackage{tikz}

\usepackage[noend]{algpseudocode}
\algdef{SE}{Begin}{End}{\textbf{begin}}{\textbf{end}}

\usepackage{lastpage}
\def\ps@jmlrtps{\let\@mkboth\@gobbletwo\def\@oddhead{}\def\@oddfoot{}\def\@evenhead{}\def\@evenfoot{}}
\begin{document}
\title{Neighbor Embedding for High-Dimensional Sparse Poisson Data}

\author{\name Noga Mudrik \email nmudrik1@jhu.edu \\
       \addr Department of Biomedical Engineering\\
       Center for Imaging Science\\
       Kavli Neuroscience Discovery Institute\\
       Johns Hopkins University\\
       Baltimore, MD 21218, USA
       \AND
       \name Adam S.\ Charles \email adamsc@jhu.edu \\
       \addr Department of Biomedical Engineering\\
       Center for Imaging Science\\
       Kavli Neuroscience Discovery Institute\\
       Johns Hopkins University\\
       Baltimore, MD 21218, USA}
\editor{unknown editor}

\maketitle

\begin{abstract}%
Across many scientific fields, measurements often represent the number of times an event occurs. For example, a document can be represented by word occurrence counts, neural activity by spike counts per time window, or online communication by daily email counts. These measurements yield high-dimensional count data that often approximate a Poisson distribution, frequently with low rates that produce substantial sparsity and complicate downstream analysis. 
A useful approach is to embed the data into a low-dimensional space that preserves meaningful structure, commonly termed dimensionality reduction. Yet existing dimensionality reduction methods, including both linear (e.g., PCA) and nonlinear approaches (e.g., t-SNE), often assume continuous Euclidean geometry, thereby misaligning with the discrete, sparse nature of low-rate count data.
Here, we propose \textit{p}-SNE (Poisson Stochastic Neighbor Embedding), a nonlinear neighbor embedding method designed around the Poisson structure of count data, using KL divergence between Poisson distributions to measure pairwise dissimilarity and Hellinger distance to optimize the embedding.
We test \textit{p}-SNE on synthetic Poisson data and demonstrate its ability to recover meaningful structure in real-world count datasets, including  weekday patterns in email communication, research area clusters in OpenReview papers, and temporal drift and stimulus gradients in neural spike recordings.
\end{abstract}
\section{Introduction}

Across many scientific domains, measurements are both high-dimensional and discrete: neuroscience experiments often analyze the number of action potentials fired by hundreds of neurons across trials~\citep{kass2005statistical}; gene expression assays measure transcript counts per cell across thousands of genes~\citep{wang2014gene}; text corpora represent documents as word occurrence counts over large vocabularies~\citep{baron2009word}. The high dimensionality of such data makes relationships between samples difficult to visualize or reason about directly. At the same time, the count nature of the observations introduces distinct statistical properties: values are non-negative integers, their variability typically scales with their mean rather than following the constant-variance assumption of Gaussian distributions, and in low-count regimes observations are characterized by an excess of zeros.

A useful approach to address the dimensionality challenge is to embed the data into a low-dimensional space that preserves its key properties; often termed dimensionality reduction. Linear methods such as principal component analysis (PCA) are limited to capturing linear relationships and miss nonlinear structure such as curved manifolds. Nonlinear neighbor embedding methods such as t-distributed stochastic neighbor embedding  (t-SNE,~\cite{van2008visualizing}) and  Uniform Manifold Approximation and Projection  (UMAP,~\cite{mcinnes2018umap}) overcome this by preserving local neighborhood structure, making them well suited for data with complex geometry. t-SNE and UMAP are now among the most widely used tools for visualizing high-dimensional data across scientific domains~\citep{moon2019visualizing, macosko2015highly}.
However, these methods assume continuous, Gaussian-distributed features. For sparse count data, these assumptions misrepresent pairwise relationships by ignoring the mean-variance coupling and the prevalence of zeros, calling for nonlinear dimensionality reduction methods that respect Poisson statistics.

Methods closer to ours  include Zero Inflated Factor Analysis (ZIFA), which accounts for zero-inflation through a latent factor model, and Potential of Heat-diffusion for Affinity-based Transition Embedding (PHATE), which captures geometric structure via diffusion distances; yet they neither directly model the Poisson likelihood in the construction of pairwise similarities nor incorporate it into the optimization cost. 
Other approaches that leverage Poisson likelihood for dimensionality reduction include GLM-PCA~\citep{townes2019feature}, Poisson GPFA~\citep{duncker2018temporal,zhao2017variational}, and Poisson matrix factorization, which are fundamentally linear and cannot capture nonlinear manifold structures, and deep generative models, e.g., single-cell variational inference (scVI~\citep{lopez2018deep}) and 
Poisson VAE 
(P-VAE~\citep{vafaii2024poisson}), which require large training datasets and are computationally expensive.

Here, we propose \textit{p}-SNE,
a nonlinear neighbor embedding method designed for high-dimensional count data. \textit{p}-SNE constructs pairwise similarities based on Poisson Kullback-Leibler (KL) divergence and optimizes the embedding via the Hellinger distance, a symmetric, bounded cost that is robust to near-zero similarities.
Our contributions include:

\begin{itemize}
    \item 
    We derive \textit{p}-SNE and describe the algorithm to determine its embedding.

  \item 
We validate \textit{p}-SNE on {simulated} Poisson data and demonstrate that it recovers ground-truth cluster structure more accurately than alternative methods.

  \item 
We demonstrate \textit{p}-SNE on three real-world count datasets spanning electronic communications, scientific text, and neural recordings, showing consistent improvements in downstream clustering and classification performance.

\end{itemize}

\section{Related Work}

Dimensionality reduction methods have been extensively developed to extract low-dimensional structure from high-dimensional data, and broadly divide into linear and nonlinear approaches, each with distinct assumptions and trade-offs.

Linear methods find a low-dimensional projection of the data as a linear combination of the original features. Principal component analysis (PCA), one of the most commonly-used methods, identifies directions of maximal variance, whereas non-negative matrix factorization (NMF)~\citep{lee1999learning} adds a non-negativity constraint.
Other methods include independent component analysis (ICA)~\citep{hyvarinen2000independent}, which seeks statistically independent components, as well as recent extensions of matrix factorization to more complex multi-array structures~\citep{mudrik2026multi, mudrik2024sibblings}.
More broadly, \cite{cunningham2015linear} showed that many of these methods can be unified as optimization problems over matrix manifolds. This shared linear structure, however, is also a shared limitation: the embedding is constrained to be a linear function of the input, which may fail to capture nonlinear structure present in complex data.

Nonlinear methods relax this constraint by constructing embeddings that preserve local neighborhood structure rather than global linear projections.
t-SNE~\citep{van2008visualizing} 
converts pairwise Euclidean  distances into conditional probabilities and minimizes the KL divergence between high- and low-dimensional distributions, producing embeddings that reveal cluster structure. UMAP~\citep{mcinnes2018umap} offers a related but faster approach grounded in Riemannian geometry and fuzzy topology. Other recent methods including  Large-scale Dimensionality Reduction Using Triplets (TriMap,~\cite{amid2019trimap}) and Pairwise Controlled Manifold Approximation (PaCMAP,~\cite{wang2021understanding})  address known limitations of t-SNE and UMAP in preserving global structure alongside local neighborhoods.
More advanced methods include  PHATE \citep{moon2019visualizing} that captures continuous transitions via graph diffusion distances, and its extensions M-PHATE \citep{gigante2019visualizing} and MM-PHATE \citep{xie2024multiway} that incorporate temporal structure to visualize evolving neural network representations. While these methods enrich the embedding with additional structure, none account for the distributional properties of count data. Consistent Embeddings of high-dimensional Recordings using Auxiliary variables (CEBRA,~\cite{schneider2023learnable}) extends this family to neuroscience settings by incorporating behavioral and task labels into the embedding objective. Despite their different formulations, all these methods rely on Euclidean or general continuous-valued distance measures to define neighborhood structure in the high-dimensional space. This makes them poorly suited for data whose natural geometry follows Poisson distribution,
where differences in integer counts carry very different statistical meaning than Euclidean distances.

A common practical workaround to analyze count data is to  apply a $\log(1+\bm{Y})$
transformation to the data before passing it to a standard dimensionality reduction method~\citep{luecken2019current,townes2019feature,booeshaghi2021normalization}. This pre-processing step compresses large counts and partially stabilizes variance, and remains the dominant approach in practice~\citep{luecken2019current, townes2019feature,kobak2019art}. 
However, it discards the distributional structure of count data, and under a Poisson generative model, the natural measure of similarity between two count vectors is not Euclidean distance on log-transformed values, but a divergence that respects the Poisson likelihood.

Several methods have been developed to incorporate Poisson statistics  directly 
into dimensionality reduction.
\cite{ling2022dimension} proposed a sparse-vector dissimilarity measure
 and plugged it into existing methods, further demonstrating that standard Euclidean distances are inappropriate for sparse count data.
Generalized PCA (GLM-PCA,~\cite{townes2019feature}) replaces the Gaussian likelihood implicit in PCA with a Poisson likelihood, providing a principled linear decomposition for count data.
 Poisson matrix factorization~\citep{gopalan2015scalable} models count observations as the product of two low-dimensional non-negative matrices, and ZIFA~\citep{pierson2015zifa} addresses zero-inflated count data through a latent factor model that explicitly accounts for dropout, but these methods cannot capture nonlinear embeddings. scVI \citep{lopez2018deep} uses a variational autoencoder with a count-based likelihood to learn low-dimensional representations of single-cell data, and the Poisson variational autoencoder (P-VAE)~\citep{vafaii2024poisson} adopts a deep generative model with Poisson-distributed latent variables; as deep generative models, both require large training datasets and are computationally demanding.

\section{Background} \label{sec:preliminaries}
Several divergence measures exist for comparing probability distributions. Standard distance-based measures such as Euclidean and Wasserstein operate on the geometry of the support rather than the likelihood structure, and thus cannot exploit the parametric form of specific distributional models such as the Poisson.
Other approaches, based on information-theoretic divergences, exploit the parametric form and are thereby more appropriate for likelihood-based comparisons. These include, e.g., the KL divergence and the Hellinger distance.

The \textbf{KL divergence} between two distributions $\mathcal{U}$ and $\mathcal{V}$ is defined as $\text{KL}(\mathcal{U} \| \mathcal{V}) = \sum_k \mathcal{U}(k) \log \frac{\mathcal{U}(k)}{\mathcal{V}(k)}$, and quantifies the information lost when $\mathcal{V}$ is used to approximate $\mathcal{U}$. It is non-negative, equals zero if and only if $\mathcal{U} = \mathcal{V}$, and is asymmetric. For two Poisson distributions with means $\lambda_1$ and $\lambda_2$, it takes the closed form: \begin{equation} \text{KL}(\text{Pois}(\lambda_1) \| \text{Pois}(\lambda_2)) = \lambda_1 \log\frac{\lambda_1}{\lambda_2} + \lambda_2 - \lambda_1 \label{eq:kl_poisson} \end{equation} The derivation is given in Appendix~\ref{app:kl_derivation}. However, the KL divergence is unbounded and diverges when $\mathcal{V}$ assigns zero probability to an event with nonzero probability under $\mathcal{U}$, which limits its use as an optimization objective when distributions contain many near-zero entries.
Symmetrizing the KL divergence via the Jensen-Shannon divergence produces a distance that is bounded and symmetric, yet still involves log terms that require numerical corrections when either distribution has zero entries. 

An alternative that does not involve log terms is the \textbf{Hellinger distance}, which uses square-root weighting to naturally handle zeros without numerical corrections. For two discrete probability distributions $\mathcal{U}$ and $\mathcal{V}$ defined over the same support, it is defined as: \begin{align} H(\mathcal{U}, \mathcal{V}) &= \sqrt{\frac{1}{2} \sum_{i} \left(\sqrt{\mathcal{U}_i} - \sqrt{\mathcal{V}_i}\right)^2} \label{eq:hellinger} \end{align} The Hellinger distance is symmetric, $H(\mathcal{U},\mathcal{V}) = H(\mathcal{V},\mathcal{U})$, and bounded, $H(\mathcal{U},\mathcal{V}) \in [0,1]$. It equals zero if and only if $\mathcal{U} = \mathcal{V}$, and equals one when the two distributions have disjoint support. Unlike KL divergence, it does not diverge when one distribution assigns zero probability to an event that the other does not, making it robust in sparse settings. Its square-root weighting also makes it smooth and differentiable, well-suited for gradient-based optimization.

\section{Model and Fitting}

\subsection{Problem Formulation}
\label{sec:problem}

Let $\bm{Y} \in \mathbb{Z}_{\geq 0}^{N \times M}$ 
denote a high-dimensional count matrix,
where $N$ is the number of samples and $M$ is the number of features. Each row $\bm{y}_n \in \mathbb{Z}_{\geq 0}^{M}$ represents a single observation, and each entry ${y}_{n,m}$ is a non-negative integer count. We assume that counts are generated under a Poisson process, so that $y_{n,m} \sim \text{Poisson}(\lambda_{n,m})$ for latent rates $\lambda_{n,m} > 0$.

\begin{figure}[t]
    \centering
    \includegraphics[width=\textwidth]{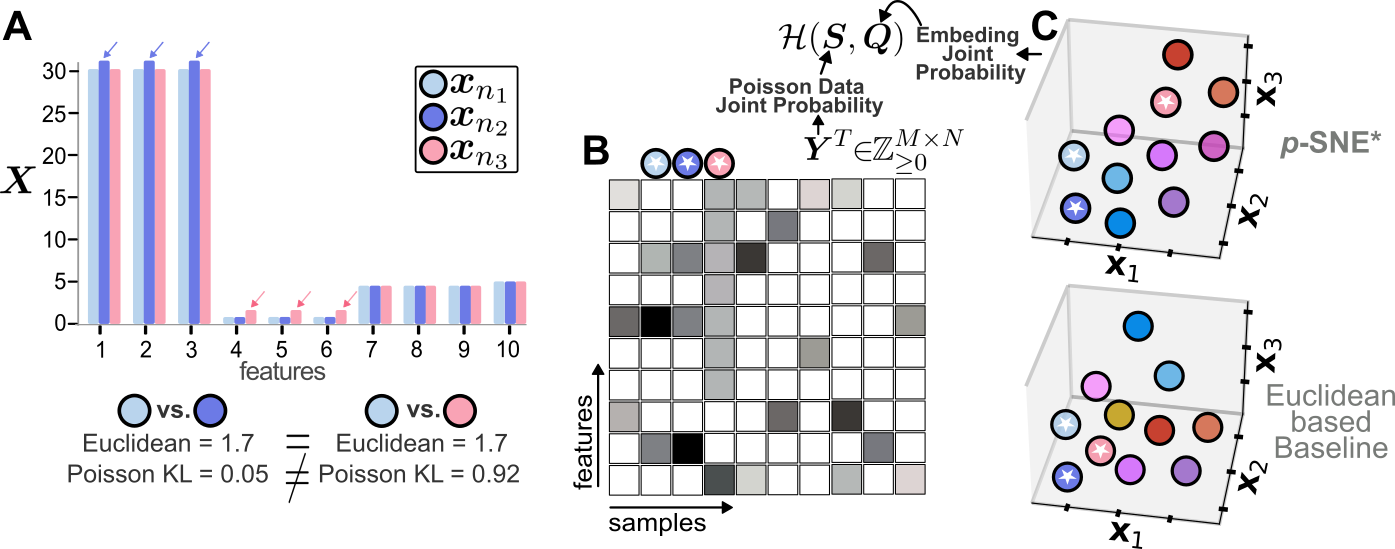}
    \caption{\textbf{Overview of \textit{p}-SNE.} \textbf{(A)}~Euclidean distance assigns the same value (1.73) to both pairs ($\bm{x}_{n_1}$ vs.\ $\bm{x}_{n_2}$ and $\bm{x}_{n_1}$ vs.\ $\bm{x}_{n_3}$), while Poisson KL divergence distinguishes them (0.05 vs.\ 0.92), since differences at low counts (pink arrows) alter the implied Poisson distribution more than equal-sized differences at high counts (blue arrows).\textbf{(B)}~\textit{p}-SNE computes pairwise Poisson KL dissimilarities from a count matrix $\bm{Y} \in \mathbb{Z}_{\geq 0}^{N \times M}$, constructs a similarity distribution $\bm{S}$, and minimizes the Hellinger cost $\mathcal{H}(\bm{S}, \bm{Q})$ over a Student-t embedding $\bm{X}$. \textbf{(C)}~\textit{p}-SNE (top) preserves the Poisson structure, keeping $\bm{x}_{n_1}$  and $\bm{x}_{n_2}$ close while separating $\bm{x}_{n_3}$, whereas the Euclidean-based baseline (bottom) fails to separate them. White stars in (B) and (C) mark the three samples from (A).}
    \label{fig:overview}
\end{figure}

Such data carries statistical properties, including non-negativity, mean-variance dependence, and, at low rates, sparsity, that existing dimensionality reduction methods that rely on Euclidean distances and Gaussian-based similarities
are not designed to handle.
As a result, applying commonly used methods to low-rate count data risks distorting pairwise structure and obscuring meaningful patterns in the embedding (Fig.~\ref{fig:overview}A).
Our goal is to find a continuous low-dimensional embedding $\bm{X} \in \mathbb{R}^{N \times P}$, where $P \ll M$, that preserves the neighborhood structure of the samples while respecting their Poisson statistics.

\subsection{Our Approach}
\label{sec:pSNE}
p-SNE seeks a low-dimensional embedding $\bm{X} \in \mathbb{R}^{N \times P}$ that faithfully captures the neighborhood structure of the high-dimensional count matrix $\bm{Y} \in \mathbb{Z}_{\geq 0}^{N \times M}$. To this end, p-SNE constructs a Poisson-based  joint probability distribution  over sample pairs in the high-dimensional space, which we denote by $\bm{S}\in\mathbb{R}^{N\times N}$, a corresponding distribution in the low-dimensional space, denoted by $\bm{Q}\in\mathbb{R}^{N\times N}$, and then optimizes $\bm{X}$ so that $\bm{S}$ and $\bm{Q}$ are similar to each other
(Fig.~\ref{fig:overview}B).
Each of these three components is designed to respect the Poisson structure of the data: $\bm{S}$ is built from KL 
divergences between the Poisson distributions implied by each pair of samples, and the discrepancy between $\bm{S}$ and $\bm{Q}$ is measured by the Hellinger distance.
p-SNE proceeds in three steps. Steps 1 and 2 are computed once from the data, while step 3 is iterative:
\begin{itemize}
    \item \textbf{Step 1:} Compute pairwise Poisson dissimilarities, denoted  as $\bm{D}\in\mathbb{R}^{N\times N}$, from $\bm{Y}$
    \item \textbf{Step 2:} Use $\bm{D}$ to construct the high-dimensional joint probability distribution $\bm{S}\in\mathbb{R}^{N\times N}$. 
    \item \textbf{Step 3:}  Initialize the latent embedding $\bm{X}$ randomly and iteratively minimize the Hellinger distance between $\bm{S}$ and $\bm{Q}$ (the similarity distribution induced by $\bm{X}$)  
\end{itemize}

\vspace{5pt}
\noindent
\textbf{Step 1: Computing pairwise dissimilarity with the KL divergence of Poissons. }
For two samples $\bm{y}_{n_1}$ and $\bm{y}_{n_2}$, we measure their per-feature dissimilarity as the KL divergence between Poisson distributions with means $y_{n_1,m}$ and $y_{n_2,m}$, i.e., $\text{KL}(\text{Pois}(y_{n_1,m}) \| \text{Pois}(y_{n_2,m}))$:
\begin{equation}
    d_{m,(n_1,n_2)} = y_{n_1,m} \log \left( \frac{y_{n_1,m} + \epsilon}{y_{n_2,m} + \epsilon} \right)+ y_{n_2,m} - y_{n_1,m}
    \label{eq:per_feature_div}
\end{equation}
where $\epsilon>0$ is a small scalar added for preventing division by zero inside the log and numerical instability.
The full pairwise dissimilarity between two samples is the sum of per-feature divergences over all $M$ features:
\begin{equation}
    D_{n_1,n_2} = \sum_{m=1}^{M} d_{m,(n_1,n_2)}
    \label{eq:full_dist}
\end{equation}
Note that since KL divergence is asymmetric, $D_{n_1,n_2} \neq D_{n_2,n_1}$ in general. 

\vspace{5pt}
\noindent
\textbf{Step 2: Calculating the high dimensional conditional probabilities based on the dissimilarity metric. }
Next, we note that the dissimilarity matrix $\mathbf{D}$ is asymmetric, which means it does not define a unique notion of neighborhood: sample $n_1$ may regard $n_2$ as a close neighbor while $n_2$ does not regard $n_1$ as close. We thus construct, using $\mathbf{D}$, a symmetric joint probability distribution $\mathbf{S}$ over sample pairs that reflects neighborhood structure from both directions.
We first obtain conditional probabilities of observing sample $n_2$ in the neighborhood of sample $n_1$  via softmax over the dissimilarities:

\begin{equation}
    p_{n_2|n_1} = \frac{\exp(-w \cdot D_{n_1,n_2})}{\sum_{\tau \neq n_1} \exp(-w \cdot D_{n_1,\tau})}
    \label{eq:conditional}
\end{equation}
where $w > 0$ controls the sharpness of the distribution, with larger $w$ concentrating probability mass on the nearest neighbors. Because $\bm{D}$ is asymmetric (Step 1), the conditionals are also asymmetric: $p_{n_2|n_1} \neq p_{n_1|n_2}$ in general. To obtain a symmetric joint distribution that incorporates both directions, we average the two conditionals:
\begin{equation}
    S_{n_1,n_2} = \frac{p_{n_2|n_1} + p_{n_1|n_2}}{2N}
    \label{eq:S}
\end{equation}
and the division by $2N$ ensures $\sum_{n_1,n_2} S_{n_1,n_2} = 1$, so $\bm{S}$ is a proper probability distribution. The matrix $\bm{S}$ is computed once and held fixed throughout the iterative model fitting.

\vspace{5pt}
\noindent
\textbf{Step 3: Cost function specification for optimizing the low-dimensional representation. }
The Poisson structure of the data is captured by the high-dimensional distribution $\bm{S}$. 
Since $\bm{X}$ is continuous and real-valued, we follow t-SNE and define a joint probability distribution $\bm{Q}$ over sample pairs using a Student-t kernel with one degree of freedom:

\begin{equation}
    k(n_1, n_2) = \left(1 + \|\bm{x}_{n_1} - \bm{x}_{n_2}\|^2\right)^{-1}
    \label{eq:student_t}
\end{equation}
The heavy tails of this kernel help alleviate the crowding problem that arises when mapping high-dimensional neighborhoods into low-dimensional space. Normalizing over all pairs yields a valid joint probability distribution:
\begin{equation}
    Q_{n_1,n_2} = \frac{k(n_1,n_2)}{\sum_{s_1 \neq s_2} k(s_1,s_2)} = \frac{\left(1 + \|\bm{x}_{n_1} - \bm{x}_{n_2}\|^2\right)^{-1}}{\sum_{s_1 \neq s_2}\left(1 + \|\bm{x}_{s_1} - \bm{x}_{s_2}\|^2\right)^{-1}}
    \label{eq:Q}
\end{equation}
The goal is to find $\bm{X}$ such that $\bm{Q}$ matches $\bm{S}$. Applying the Hellinger distance from Eq.~\ref{eq:hellinger} to $\bm{S}$ and $\bm{Q}$, the cost function minimized by p-SNE is:
\begin{equation}
    \mathcal{L} = \mathcal{H}(\bm{S}, \bm{Q}) = \sqrt{\frac{1}{2} \sum_{n_1, n_2}\left(\sqrt{S_{n_1,n_2}} - \sqrt{Q_{n_1,n_2}}\right)^2}
    \label{eq:cost}
\end{equation}

\begin{algorithm}[ht]
\caption{\textit{p}-SNE}
\label{alg:pSNE}
\begin{algorithmic}[1]
\State \textbf{Input:} Count matrix $\bm{Y} \in \mathbb{Z}_{\geq 0}^{N \times M}$, embedding dimension $P$, sharpness $w$, learning rate $\eta$, momentum $\mu$, exaggeration factor $\alpha$, exaggeration iterations $K_{\text{ex}}$, max iterations $K$, tolerance $\tau$, stability constant $\epsilon$
\State \textbf{Output:} Embedding $\bm{X} \in \mathbb{R}^{N \times P}$
\State \textbf{Step 1:} Compute $D_{n_1,n_2} = \sum_{m=1}^{M} d_{m,(n_1,n_2)}$ for all pairs \hfill $\triangleright$ Eq.~\ref{eq:per_feature_div},~\ref{eq:full_dist}
\State \textbf{Step 2:} Compute conditionals $p_{n_2|n_1}$ via softmax over $\bm{D}$ with sharpness $w$ \hfill $\triangleright$ Eq.~\ref{eq:conditional}
\State Symmetrize: $S_{n_1,n_2} = (p_{n_2|n_1} + p_{n_1|n_2}) / 2N$ \hfill $\triangleright$ Eq.~\ref{eq:S}
\State \textbf{Step 3:} Initialize $\bm{X}$ from Student-t($\nu=3$), set $\bm{v} \leftarrow \bm{0}$
\For{$k = 1, \ldots, K$}
    \If{$k \leq K_{\text{ex}}$}
        \State $\bm{S}_{\text{eff}} \leftarrow \alpha \bm{S} \,/\, \sum \alpha \bm{S}$ \hfill $\triangleright$ Early exaggeration
    \Else
        \State $\bm{S}_{\text{eff}} \leftarrow \bm{S}$
    \EndIf
    \State Compute $\bm{Q}$ from $\bm{X}$ using Student-t kernel \hfill $\triangleright$ Eq.~\ref{eq:Q}
    \State Compute $\mathcal{L} = \mathcal{H}(\bm{S}_{\text{eff}}, \bm{Q})$ \hfill $\triangleright$ Eq.~\ref{eq:cost}
    \If{$|\mathcal{L}^{(k)} - \mathcal{L}^{(k-1)}| < \tau$}
        \State \textbf{break}
    \EndIf
    \State $\bm{v} \leftarrow \mu \bm{v} - \eta \,\partial \mathcal{L} / \partial \bm{X}$ \hfill $\triangleright$ Eq.~\ref{eq:momentum}
    \State $\bm{X} \leftarrow \bm{X} + \bm{v}$ \hfill $\triangleright$ Eq.~\ref{eq:update}
\EndFor
\State \Return $\bm{X}$
\end{algorithmic}
\end{algorithm}

The choice of divergence differs between these stages. The KL divergence is the natural dissimilarity for comparing Poisson observations: it respects the parametric form of the Poisson likelihood and captures the mean-variance coupling inherent to count data. However, as a cost function comparing $\bm{S}$ and $\bm{Q}$, KL divergence is asymmetric, unbounded, and assigns infinite cost when one distribution places zero probability where the other does not, a situation that arises frequently when similarities are sparse. The Hellinger distance, by contrast, is symmetric, bounded, and remains well-behaved at zero entries, making it better suited as the optimization objective.

\vspace{5pt}
\noindent
\textbf{Cost function optimization.}
After $\bm{S}$ is computed (step 2), the embedding $\bm{X}$ is initialized from a Student-t distribution with $\nu=3$ degrees of freedom and updated by gradient descent with momentum (derivation in App.~\ref{app:gradients}). The update rule at iteration $k$ with velocity $\bm{v}$ is:
\begin{equation}
    \bm{v}^{(k+1)} = \mu \bm{v}^{(k)} - \eta \frac{\partial  \mathcal{L} }{\partial \bm{X}^{(k)}}
    \label{eq:momentum}
\end{equation}
\begin{equation}
    \bm{X}^{(k+1)} = \bm{X}^{(k)} + \bm{v}^{(k+1)}
    \label{eq:update}
\end{equation}
where $\eta > 0$ is the learning rate and $\mu \in [0,1)$ is the momentum coefficient. Early exaggeration~\citep{van2008visualizing}, a technique that temporarily multiplies $\bm{S}$ by a constant factor greater than one and renormalizes, is applied during the initial iterations to encourage the formation of well-separated clusters before fine-grained structure is resolved. The optimization terminates when the absolute change in cost falls below a tolerance parameter $\tau$ (App.~\ref{app:tau}).

\section{Experiments}
We evaluate \textit{p}-SNE on two synthetic 
and three 
real-world datasets. 
The synthetic data experiments  
use Poisson-distributed generated data with known ground 
truth groups, which enables controlled evaluation of whether \textit{p}-SNE recovers the true underlying structure. The three real-world datasets span distinct  domains in which count data arises naturally: email communication records, academic paper word counts from OpenReview, and neural spike counts from the Allen Brain Observatory. In each case, we compare \textit{p}-SNE against nine baselines, including general-purpose dimensionality reduction methods (t-SNE, UMAP,  Isomap, Multidimensional scaling (MDS), Spectral, NMF) and methods adapted for count data, including  $t$-SNE with variance-stabilizing normalization on the data ($t$-SNE-log and  $t$-SNE-$\sqrt{\cdot}$), and~\cite{ling2022dimension}.
Data pre-processing details are provided in Appendix~\ref{app:preprocessing}.
\subsection{Synthetic Data}
\begin{figure}[t]
    \centering
\includegraphics[width=0.95\linewidth]{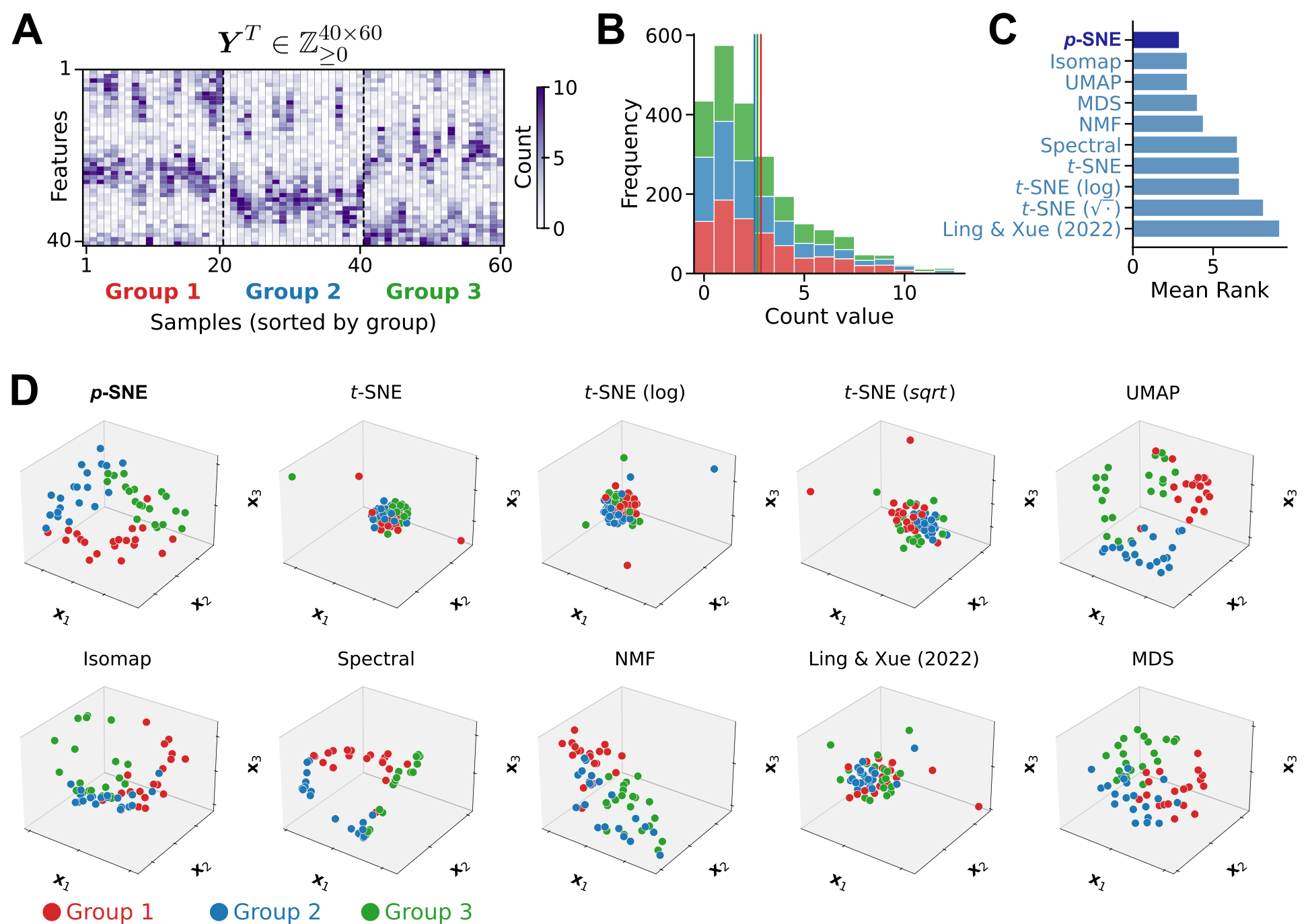}
    \caption{\textbf{Angular Embedding: \textit{p}-SNE recovers group structure on a Poisson count benchmark.}
    \textbf{(A)}~Raw count matrix $\bm{Y} \in \mathbb{Z}_{\geq 0}^{60 \times 40}$, sorted by group label.
    \textbf{(B)}~Count distribution per group; vertical lines indicate per-group means.
    \textbf{(C)}~Mean rank across classification metrics (kNN, SVM, k-means ARI) for \textit{p}-SNE and nine baselines; lower rank is better.
    \textbf{(D)}~3D embeddings colored by group label. \textit{p}-SNE ($w$=1.0) achieves the best mean rank and the clearest visual cluster separation among all methods.}
    \label{fig:synth_angular}
\end{figure}

\begin{figure}[t]
    \centering
    \includegraphics[width=\textwidth]{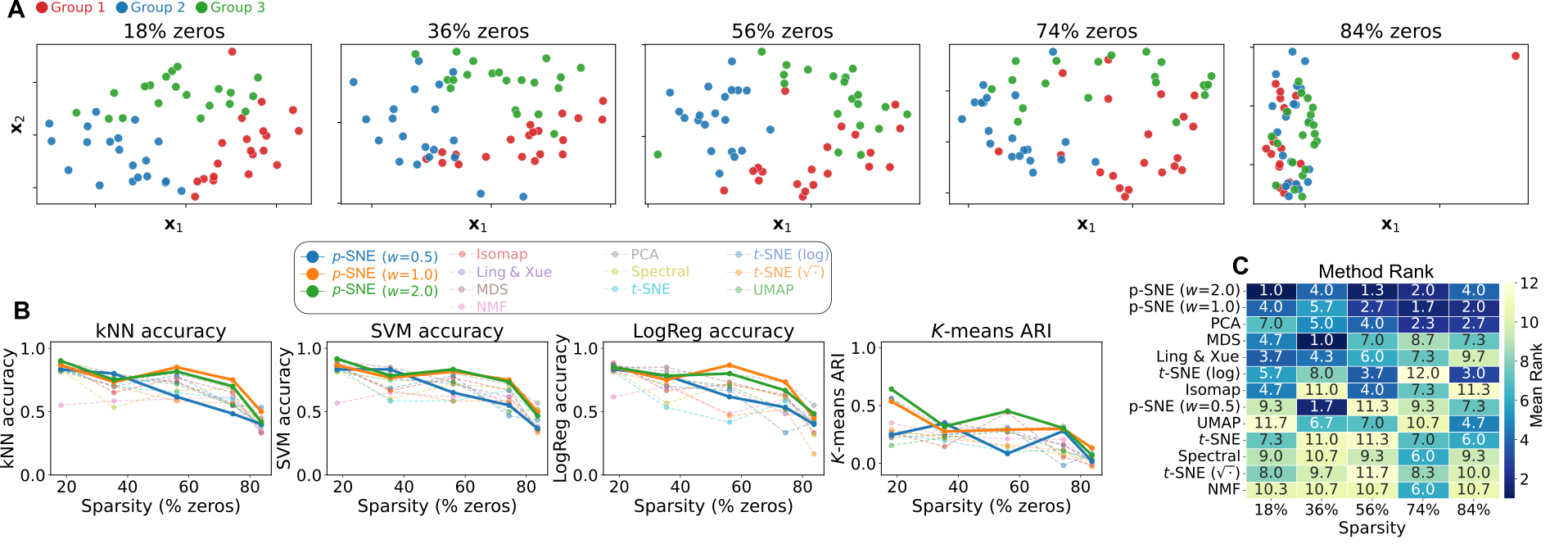}
\caption{\textbf{Effect of sparsity on embedding quality (Angular Embedding, $M = 40$ features).}
\textbf{(A)}~\textit{p}-SNE ($w$=1.0) 2D embeddings at five sparsity levels, generated by rescaling the Poisson rate parameters. Group separation degrades gradually but remains visible up to 74\% zeros.
\textbf{(B)}~Classification and clustering metrics as a function of sparsity for \textit{p}-SNE ($w \in \{0.5, 1.0, 2.0\}$) and ten baselines, evaluated on the 2D embeddings. \textit{p}-SNE maintains higher kNN, SVM, and logistic regression accuracy than all baselines across most sparsity levels, with the largest advantage at 56-74\% zeros. At extreme sparsity (84\%), all methods degrade substantially.
\textbf{(C)}~Mean rank across kNN, SVM, and $K$-means ARI (lower is better). \textit{p}-SNE variants consistently occupy the top ranks, while classical methods (PCA, MDS, NMF) deteriorate most rapidly with increasing sparsity.}
\label{fig:sparsity_sweep}

\end{figure}

We first evaluate \textit{p}-SNE on two synthetic datasets in which samples are drawn from a Poisson distribution with known structure embedded on a nonlinear manifold. In the first experiment (\textbf{Angular Embedding}), three groups of 20 samples each are generated with $M=40$ features, arranged along a nonlinear manifold with angular group structure and Poisson rates ranging from 1.0 to 9.0 a.u. In the second experiment (\textbf{Sparse Sequential Embedding}), four groups of 30 samples each are generated with $M=30$ features along a continuous nonlinear manifold, with substantially lower rates (0.1 to 2.6 a.u.) that result in high data sparsity. For each dataset, we compare \textit{p}-SNE against the nine baselines described above.

\paragraph{Angular Embedding.}
We simulate a neural population recording in which each feature represents a neuron and each sample represents a trial, yielding a count matrix $\bm{Y} \in \mathbb{Z}_{\geq 0}^{60 \times 40}$ with three trial groups (Fig.~\ref{fig:synth_angular}A). Each feature has a preferred position on the underlying manifold, and $y_{n,m}$ is a Poisson draw whose mean decreases with the difference in manifold coordinates between the stimulus and the feature's preferred position (details in App.~\ref{app:data_generation_synth1}).

We apply \textit{p}-SNE and the nine baselines to $\bm{Y}$ and compare their 3D embeddings (Fig.~\ref{fig:synth_angular}D). \textit{p}-SNE recovers an embedding with separation of the three groups, while the nine baselines produce collapsed or substantially more mixed embeddings. 
In particular, t-SNE and its variants with log and square-root, as well as~\citep{ling2022dimension} collapse into a single mixed blob affected by outliers, as the zoomed-in view in Fig.~\ref{fig:app_synth_standard_zoomed} also confirms. Isomap, NMF, and MDS achieve partial but inadequate separation, with significant boundary overlap between groups.  UMAP and Spectral Embedding overcompensate by introducing artificial fragmentation, incorrectly splitting a group (Group 3 and Group 2, respectively) into disjointed sub-clusters. In contrast, \textit{p}-SNE maintains a clearer separation of the three groups. 
We then evaluated quantitatively how well the groups are separated in the low-dimensional space, and ranked the methods across \textit{k}-nearest neighbors (kNN), Support Vector Machine (SVM), and \textit{k}-means Adjusted Rand Index (ARI) metrics that quantify group separation linearly and nonlinearly; \textit{p}-SNE achieves the best mean rank (Fig.~\ref{fig:synth_angular}C). 

Notably, when coloring the embeddings by the stimulus angular position (Fig.~\ref{fig:synth_angular_by_t}), \textit{p}-SNE preserves the angular ordering, while the other methods either collapse the angular structure or split stimuli with similar angular positions into separate clusters. 
Similar trends appear when using 2 embedding dimensions (Fig.~\ref{fig:app_synth_standard_2d}), when varying dimensionality across $d \in \{2, 3, 5, 10, 15, 20\}$ (Fig.~\ref{fig:ndim_standard}), and when sweeping \textit{p}-SNE's parameters $w$ and $\eta$ (Fig.~\ref{fig:app_sweep_standard}).
To further assess robustness to data sparsity, we gradually increased the fraction of zeros by rescaling the Poisson rates downward (Fig.~\ref{fig:sparsity_sweep}). \textit{p}-SNE maintains the highest classification accuracy across most sparsity levels, with the advantage most pronounced at moderate-to-high sparsity (56-74\% zeros).
Embeddings for \textit{p}-SNE and four representative baselines at each sparsity level confirm that baselines progressively lose group structure while \textit{p}-SNE maintains clearer separation (Fig.~\ref{fig:sparsity_1_app_baselines}).

\begin{figure}[t]
    \centering
    \includegraphics[width=0.82\linewidth]{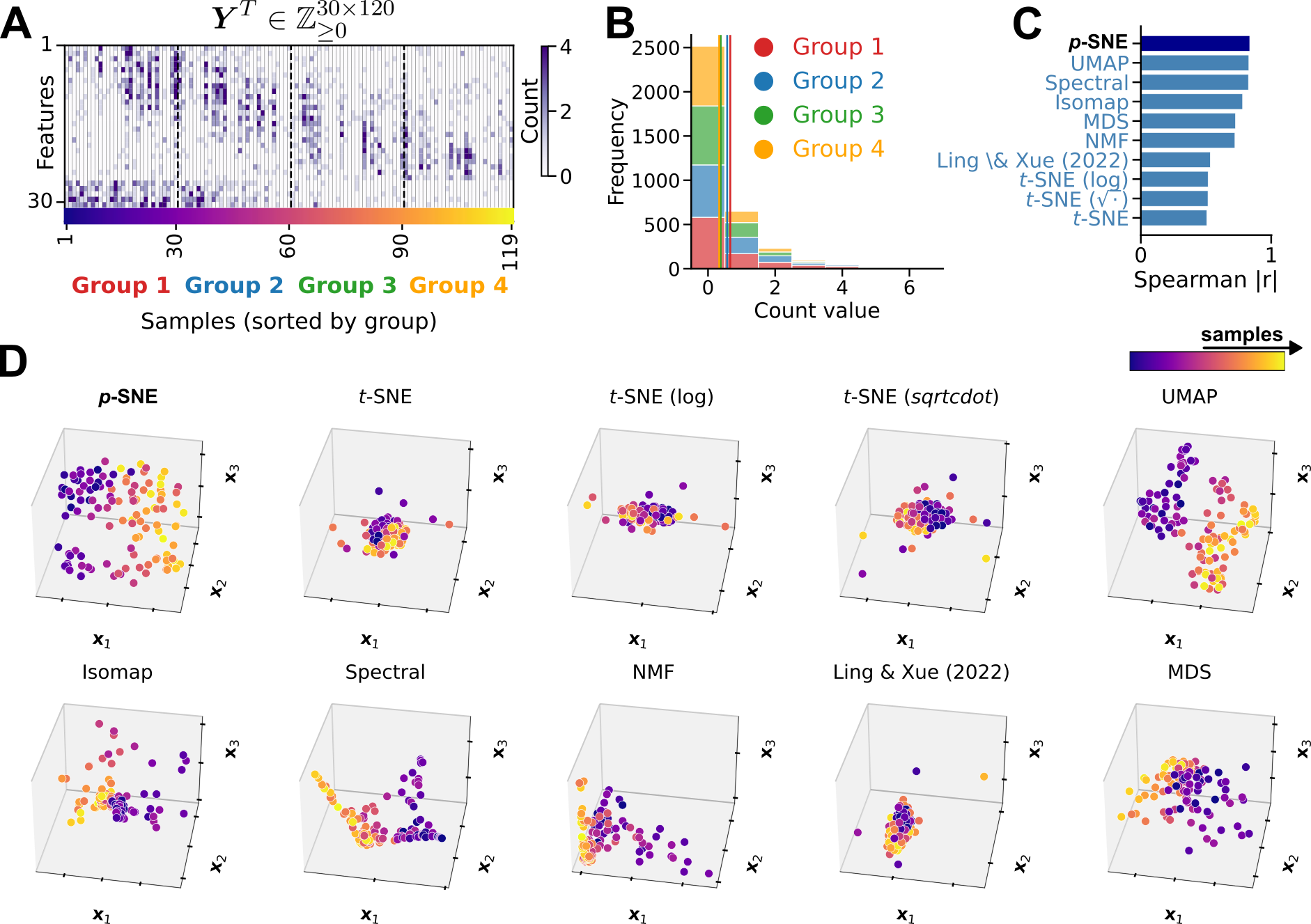}
    \caption{\textbf{Sparse Sequential Embedding: \textit{p}-SNE recovers manifold structure under high sparsity.}
    \textbf{(A)}~Raw count matrix $\bm{Y} \in \mathbb{Z}_{\geq 0}^{119 \times 30}$, sorted by ground-truth manifold parameter $t$.
    \textbf{(B)}~Count distribution per group; vertical lines indicate per-group means.
    \textbf{(C)}~Spearman correlation between embedding dimensions and $t$ for \textit{p}-SNE and nine baselines; higher is better. \textit{p}-SNE ($w$=2.0) ranks second overall, outperforming all $t$-SNE variants and Ling \& Xue (2022).
    \textbf{(D)}~3D embeddings colored by $t$ value. \textit{p}-SNE recovers a smooth continuous gradient, indicating faithful manifold recovery.}
    \label{fig:synth_sparse_linear}
\end{figure}

\paragraph{Sparse Sequential Embedding.}
The second dataset is more challenging due to substantially lower Poisson rates (0.1 to 2.6 a.u.), resulting in high sparsity and a data matrix $\bm{Y} \in \mathbb{Z}_{\geq 0}^{119 \times 30}$ (Fig.~\ref{fig:synth_sparse_linear}A,B; App.~\ref{app:data_generation_synth2}).
\textit{p}-SNE ($w$=2.0)
recovers the manifold structure more faithfully than most methods, 
as reflected in the smooth color gradient visible in the embedding (Fig.~\ref{fig:synth_sparse_linear}D). In contrast, $t$-SNE and its variants collapse into a single mixed blob dominated by outliers, while Spectral Embedding separates adjacent manifold segments into disconnected clusters. UMAP splits the samples into two groups while ignoring their internal ordering, and Isomap fails to capture the sequential progression along the manifold. Quantitative evaluation via Spearman correlation with the ground-truth manifold parameter (Fig.~\ref{fig:synth_sparse_linear}C) shows that \textit{p}-SNE ranks second overall, closely behind Spectral Embedding, and outperforms all $t$-SNE variants and~\cite{ling2022dimension}. 

\begin{figure}
    \centering
    \includegraphics[width=1\linewidth]{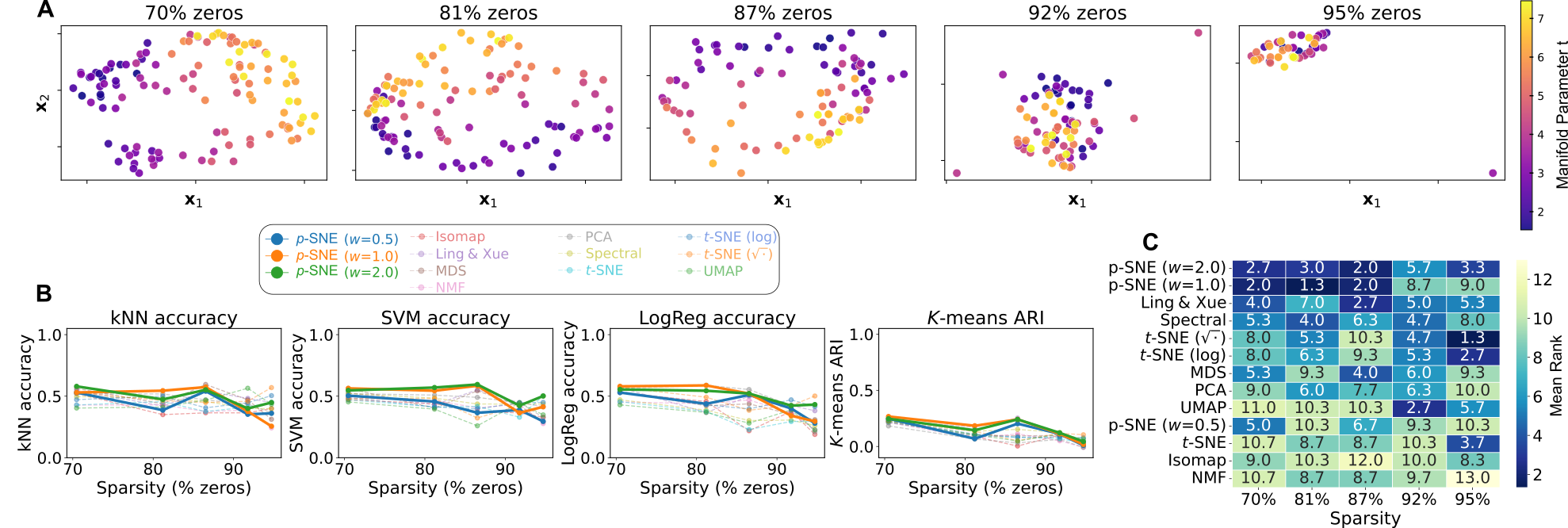}
    \caption{\textbf{Sparsity sweep on the Sparse Sequential Embedding dataset.} \textbf{(A)}~\textit{p}-SNE embeddings ($w{=}2.0$) colored by the continuous manifold parameter $t$, at five sparsity levels (70\%-95\% zeros). The smooth color gradient indicates that \textit{p}-SNE recovers the latent manifold structure even at extreme sparsity. \textbf{(B)}~Classification and clustering metrics as a function of sparsity for \textit{p}-SNE variants (solid) and baselines (dashed). \textit{p}-SNE maintains higher kNN, SVM, and logistic regression accuracy than all baselines across sparsity levels, and the gap widens at high sparsity. \textbf{(C)}~Mean rank across metrics (kNN, SVM, LogReg, $K$-means ARI) at each sparsity level (lower is better). \textit{p}-SNE ($w{=}1.0$ and $w{=}2.0$) consistently rank first or second, while baseline methods degrade in rank as sparsity increases.}
    \label{fig:sparsity_sequential}
\end{figure}

Notably, \textit{p}-SNE's recovery of the latent manifold ordering remains stable across embedding dimensionalities $d \in \{2, 3, 5, 10, 15, 20\}$ (Fig.~\ref{fig:ndim_xor}) and under varying \textit{p}-SNE parameters $w$ and $\eta$ (Fig.~\ref{fig:app_sweep_xor}).
We further evaluated robustness to increasing sparsity by rescaling the Poisson rates downward (Fig.~\ref{fig:sparsity_sequential}) 
 and compared to  baselines across sparsity levels (Figs.~\ref{fig:sparsity_1_app_baselines},~\ref{fig:sparsity_app_sequential}).
\textit{p}-SNE preserves the manifold structure across sparsity levels up to 92\% zeros,  ranking first or second among all methods.
The same trend is visible in the per-method embeddings, where \textit{p}-SNE preserves the smooth color gradient across all sparsity levels while baselines, particularly PCA, lose this structure (Fig.~\ref{fig:sparsity_app_sequential}).

Interestingly, 
when the same baselines are applied to the `noise-free' rate matrix $\bm{\Lambda}$ (Fig.~\ref{fig:rate}), several baselines recover structure comparable to what \textit{p}-SNE achieves from the noisy counts, suggesting that \textit{p}-SNE effectively recovers the underlying rate geometry despite the Poisson noise.

\subsection{Email Communication Data}
We next apply \textit{p}-SNE to a real-world count dataset derived from personal email metadata. We collected emails received in the inbox of one of the authors from October 2024 to February 2026, and constructed a count matrix where each entry $y_{nm}$ records how many emails from sender $m$ were received on day $n$. After filtering and selecting the 150 most frequent senders across 370 days, the resulting matrix is naturally sparse, with an approximate Poisson count distribution (preprocessing details in App.~\ref{app:email_preproc}).

\begin{figure}[t]
    \centering
    \includegraphics[width=0.9\linewidth]{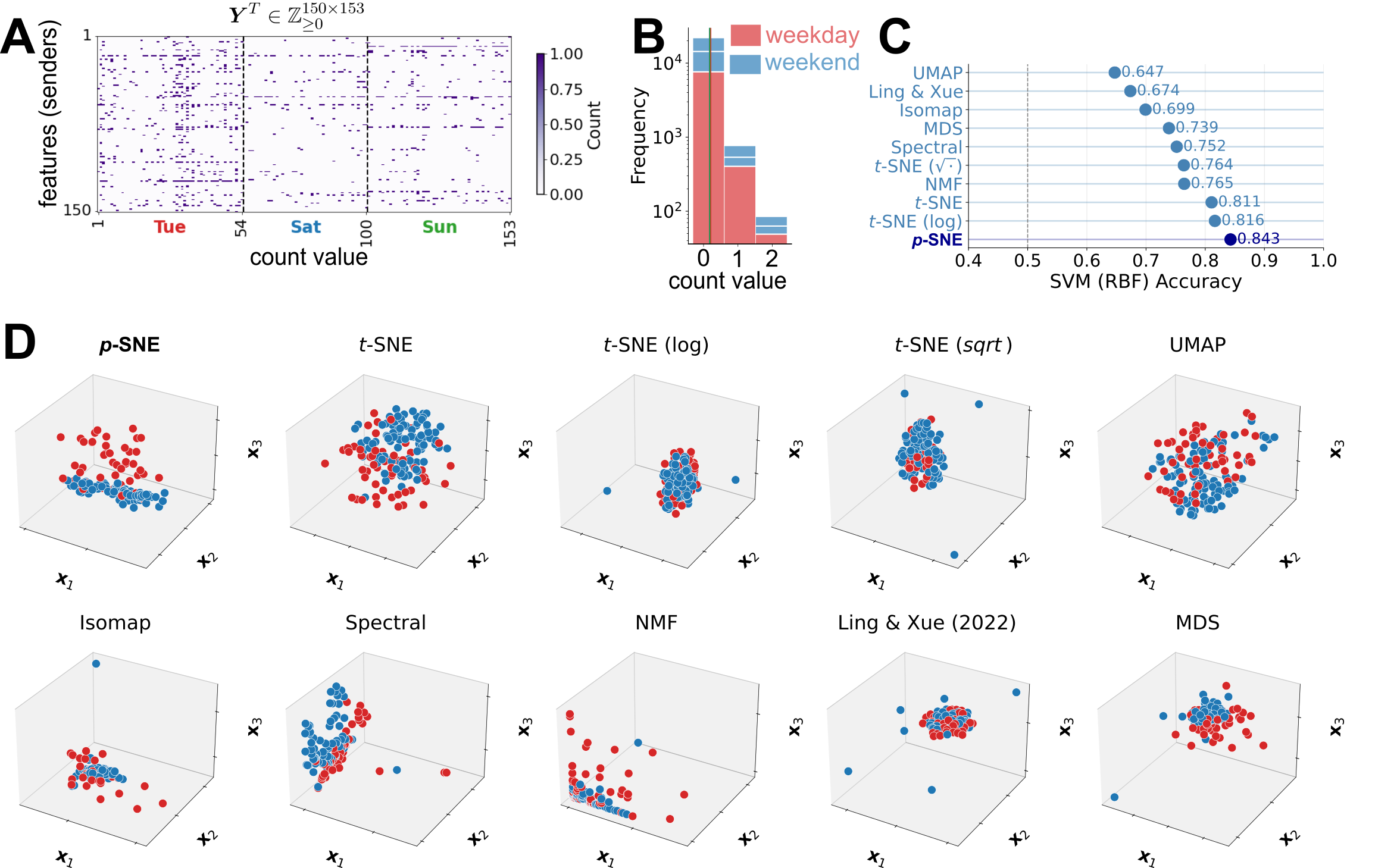}
    \caption{\textbf{Email sender count data.} \textbf{(A)}~Raw count matrix $\bm{Y} \in \mathbb{Z}_{\geq 0}^{153 \times 150}$ (days$\times$senders), sorted by label: Tuesday, Saturday, and Sunday. \textbf{(B)}~Count value distribution, stratified by weekday (red) and weekend (blue). \textbf{(C)}~SVM (RBF kernel) classification accuracy on 3D embeddings (5-fold cross-validation, for visualization here Tuesday vs.\ weekend; full data in Fig.~\ref{fig:weekend_appendix_email}). \textit{p}-SNE achieves the highest accuracy (0.843). \textbf{(D)}~3D embeddings colored by label (red = Tuesday, blue = weekend). \textit{p}-SNE produces clearer spatial separation between weekday and weekend patterns than all baselines.}
    \label{fig:email}
\end{figure}

We embedded $\bm{Y}$ with \textit{p}-SNE ($w=2.0$) and the nine baselines into three dimensions, and explored them with respect to weekday vs. weekend (Fig.~\ref{fig:email}, Fig.\ref{fig:weekend_appendix_email}).
 The baselines, including $t$-SNE, MDS, Isomap, and~\cite{ling2022dimension} struggle with complete mixing and collapsing into single dense blobs, while NMF results in smeared distributions with severe overlap. UMAP and Spectral Embedding, despite capturing some global structure or slight grouping of the classes, still suffer from significant boundary overlap and heavily mixed regions.
In contrast, \textit{p}-SNE  clearly separates between weekend and weekday emails.

To quantify the separation in the embedding space, we trained an SVM classifier (Radial Basis Function (RBF)  kernel, 5-fold cross-validation) to predict weekday versus weekend from the 3D embedding coordinates, where \textit{p}-SNE achieves the highest accuracy among all methods (Fig.~\ref{fig:email}C). We note that this classification score serves as a proxy for evaluating how well the embedding preserves group structure, not as an end goal of the dimensionality reduction method.

\subsection{OpenReview Academic Paper Abstracts}

We collected 23,944 paper abstracts from ICLR 2024, ICLR 2025, and TMLR via the OpenReview API and represented each paper as a bag-of-words count vector. To obtain clean labels, we assigned each paper to one of twelve research areas (e.g., reinforcement learning, graph neural networks, neuroscience, see App.~\ref{app:openreview}) based on domain-specific keyword matching on the raw abstract text; papers matching zero or multiple areas were excluded. 

\begin{figure}[t]
    \centering
    \includegraphics[width=1\linewidth]{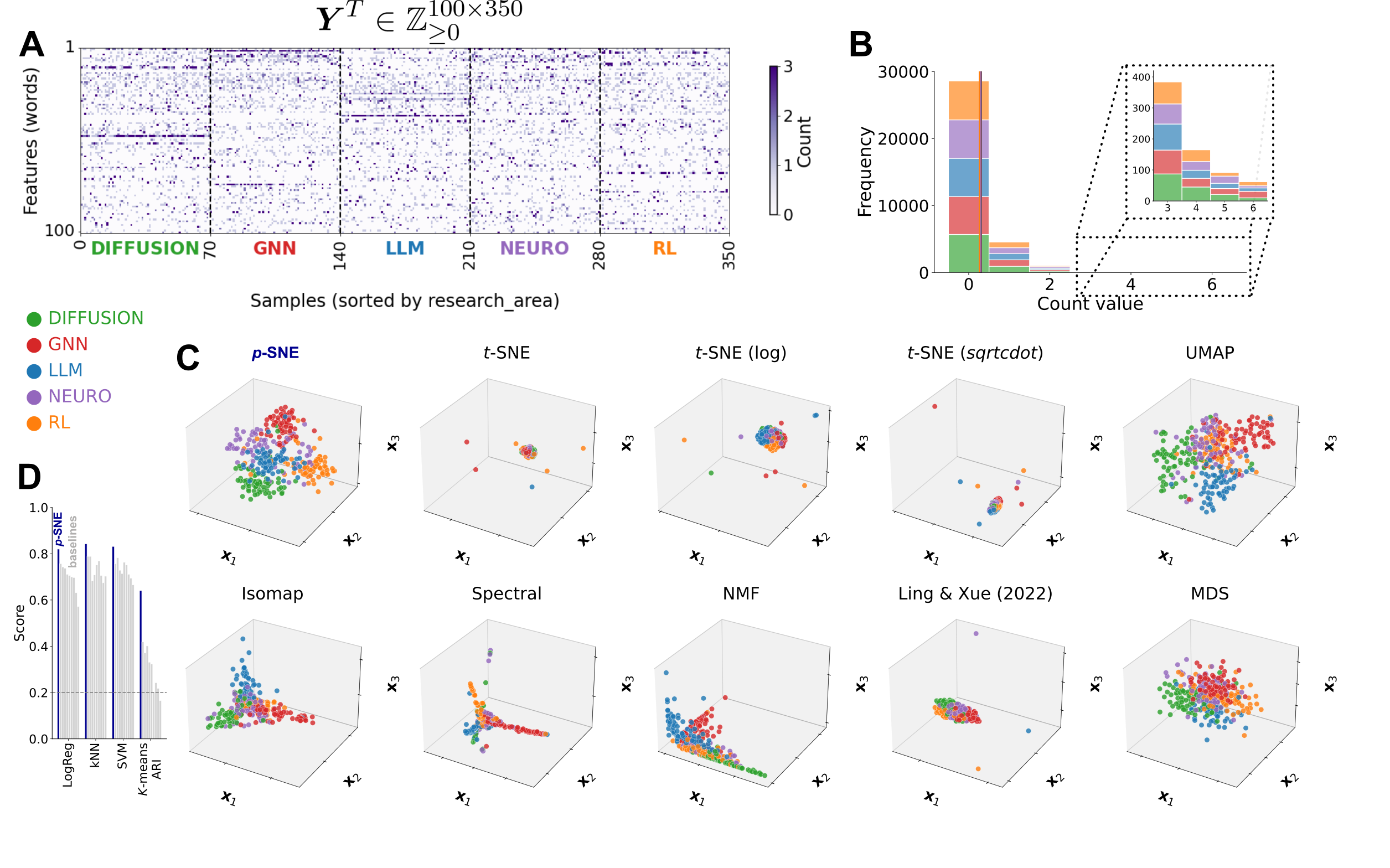}
    \caption{\textbf{Word-count experiment (OpenReview).} 350 academic papers from ICLR 2024/2025 and TMLR, represented as word-count vectors (350 paper samples (5 research areas with 70 papers each)  $\times$ 100 word features). \textbf{(A)} Raw count matrix $\mathbf{Y}$ sorted by research area, showing sparse Poisson structure with area-specific word usage patterns. \textbf{(B)} Count value distribution (81.7\% zeros, mean count 0.29). \textbf{(C)} 3D embeddings colored by research area. \textit{p}-SNE ($w$=1.0) produces the clearest cluster separation across all five areas, while baselines such as $t$-SNE collapse the structure and UMAP partially recovers it. \textbf{(D)} Classification and clustering accuracy across four metrics. \textit{p}-SNE achieves the highest score on all metrics (LogReg: 0.82, kNN: 0.84, SVM: 0.83, $K$-means ARI: 0.64), substantially outperforming all nine baselines.}
    \label{fig:openreview}
\end{figure}

We retained a random subset of 350 papers (70 per area) from five areas, removed domain keywords, prepositions, and stop words, and selected the 100 most frequent words as features (processing details in App.~\ref{app:openreview}). The resulting count matrix $\mathbf{Y} \in \mathbb{Z}_{\geq 0}^{350\times 100}$ is 81.7\% sparse and exhibits approximate Poisson statistics (mean dispersion 2.10, Fig.~\ref{fig:openreview}A-B).

We applied \textit{p}-SNE ($w{=}1.0$) and compared against the baselines. The 3D embeddings (Fig.~\ref{fig:openreview}C)  show that \textit{p}-SNE produces clearly separated clusters for all five research areas.
In contrast, other methods like $t$-SNE collapse the data into a single mass and other baselines such as UMAP and Isomap recover only partial structure with smeared distributions and significant overlap between areas.
Quantitatively (Fig.~\ref{fig:openreview}D), \textit{p}-SNE achieves the highest score both in terms of logistic regression to separate the groups (cross-validation accuracy of 0.82),  kNN (accuracy 0.84), and SVM (0.83), as well as a $K$-means ARI of 0.64, outperforming the other methods.

\subsection{Neural Recordings Data}

\begin{figure}[t]
    \centering \includegraphics[width=0.95\textwidth]{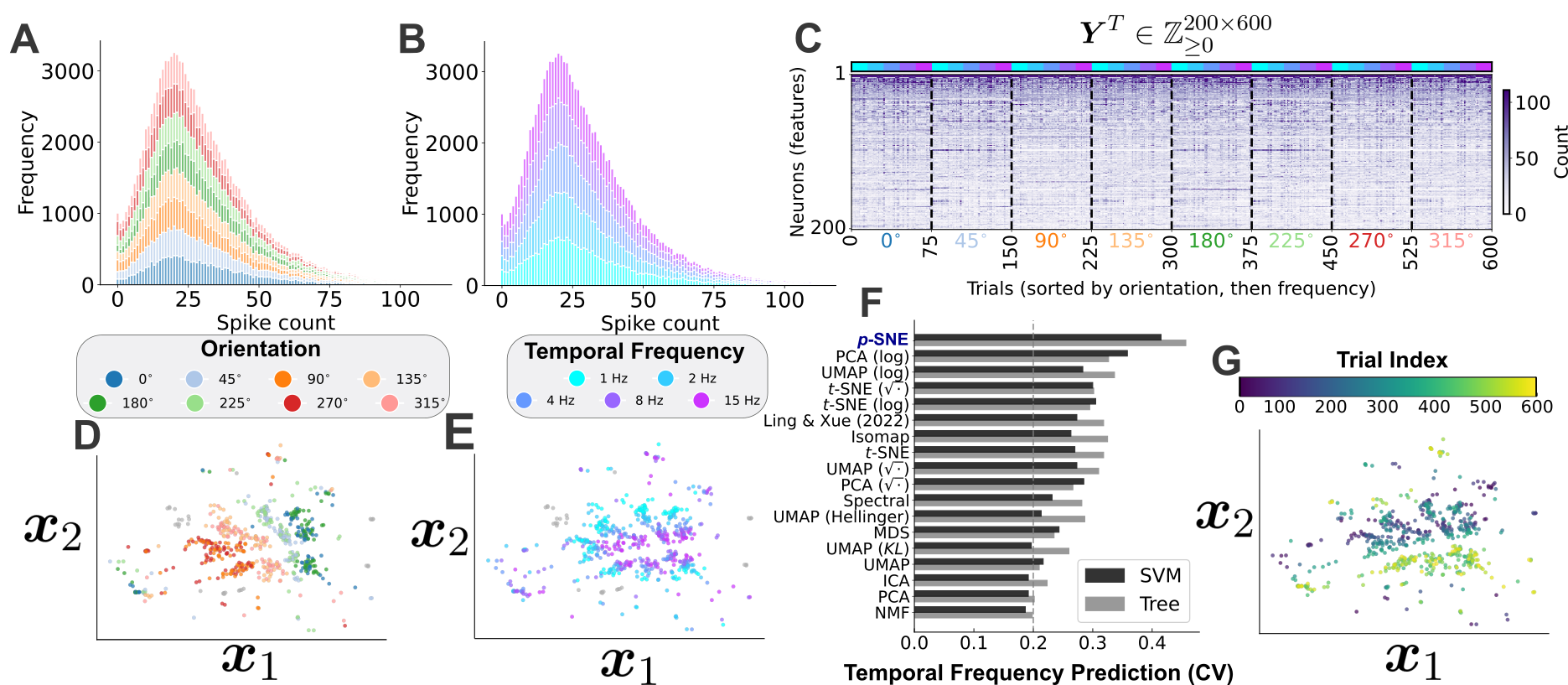} \caption{\textbf{Neuroscience experiment: neural spike count data from the Allen Brain Observatory.} \textbf{(A)}~Spike count distribution stratified by stimulus orientation (8 classes, color-coded). Distributions overlap substantially across orientations, reflecting the difficulty of the classification task. \textbf{(B)}~Spike count distribution stratified by temporal frequency (5 classes). Lower frequencies (1-2 Hz) produce slightly higher counts than higher frequencies (8-15 Hz). \textbf{(C)}~Raw count matrix $\bm{Y} \in \mathbb{Z}_{\geq 0}^{630 \times 200}$ (trials $\times$ neurons), sorted first by orientation then by temporal frequency within each orientation group (colored bar, top). Dashed lines mark orientation boundaries. \textbf{(D)}~2D \textit{p}-SNE embedding ($w{=}0.5$) colored by orientation. Several orientations form distinguishable clusters. \textbf{(E)}~Same embedding colored by temporal frequency. Low-frequency trials (cyan) separate from high-frequency trials (purple). \textbf{(F)}~Classification accuracy (5-fold cross-validation) on 2D embeddings for temporal frequency prediction. \textit{p}-SNE (dark bar) achieves the highest accuracy on both SVM and decision tree classifiers against baselines. \textbf{(G)}~Same embedding colored by trial index (temporal order within the session). A visible gradient indicates that \textit{p}-SNE preserves the temporal drift in neural activity across the recording session.} \label{fig:neuro}

\end{figure}

We evaluate \textit{p}-SNE on extracellular electrophysiology data~\citep{de2020large} from the Allen Brain Observatory Visual Coding dataset (accessed via the DANDI archive, dandiset 000021). This dataset contains large-scale Neuropixels recordings from the mouse visual system, collected as part of a standardized survey of neural activity in awake, head-fixed mice passively viewing visual stimuli. We selected a single random session
containing 1{,}891 recorded units, of which 1{,}193 were classified as ``good'' quality based on standard spike-sorting quality metrics. These units spanned visual cortex (VISp, VISam, VISrl, VISal, VISpm, VISl), hippocampus (CA1, CA3, DG), and thalamus (LGd).

During the session, the mouse was presented with drifting sinusoidal gratings comprising 8 orientations ($0^{\circ}, 45^{\circ}, \ldots, 315^{\circ}$) crossed with 5 temporal frequencies ($1, 2, 4, 8, 15$ Hz), yielding 40 unique stimulus conditions repeated approximately 15 times each, for a total of 630 trials. 
We counted the total number of spikes fired by each neuron during each 2-second trial, yielding a count matrix $\bm{Y} \in \mathbb{Z}_{\geq 0}^{630 \times 200}$ where entry $y_{n,m}$ represents the total spike count of neuron $m$ during trial $n$.
We removed silent neurons (fewer than 2 total spikes overall) and selected the 200 least active neurons.
The resulting matrix (e.g., for 200 neurons) exhibits high sparsity, with a mean spike count of 0.241 per trial and 83\% zero entries (Fig.~\ref{fig:neuro}A-C).

The \textit{p}-SNE ($w = 0.5$) embedding reveals structure along multiple axes of variation. When colored by stimulus orientation (Fig.~\ref{fig:neuro}D), several orientation classes form distinguishable spatial clusters, and when colored by temporal frequency (Fig.~\ref{fig:neuro}E), a different embedding dimension separates trials from low to high frequency. This separation is confirmed quantitatively: an SVM and a decision tree classifier trained to predict temporal frequency from the embedding coordinates show that \textit{p}-SNE achieves higher accuracy than all alternative methods (Fig.~\ref{fig:neuro}F). \textit{p}-SNE further captures temporal drift across trials, with early trials (purple and dark green, Fig.~\ref{fig:neuro}G) separated from late trials (light green), indicating that \textit{p}-SNE preserves the gradual shift in neural activity over the course of the recording session.

\section{Discussion}

Here, we presented \textit{p}-SNE, a neighbor embedding method for dimensionality reduction of sparse count data. \textit{p}-SNE constructs pairwise similarities from Poisson KL divergence and optimizes the embedding via a Hellinger distance objective, both grounded in the statistical structure of count observations. We evaluated \textit{p}-SNE on two synthetic and three real-world datasets spanning email communications, academic text, and neural recordings, and showed that it consistently recovers meaningful structure that standard baselines miss or recover less faithfully.

Across our experiments, \textit{p}-SNE's advantage over baselines is most pronounced when the data is highly sparse and counts are low. In the Sparse Sequential Embedding synthetic experiment, where Poisson rates range from 0.1 to 2.6 a.u., \textit{p}-SNE recovers the manifold structure while most Euclidean-based methods fail. In the neural data, where 83\% of entries are zero, \textit{p}-SNE achieves the highest classification accuracy for temporal frequency.

\textit{p}-SNE has several limitations. Like other neighbor embedding methods, \textit{p}-SNE requires computing a pairwise distance matrix, which scales as $\mathcal{O}(N^2 M)$ and can become  computationally demanding for datasets with thousands of samples. This bottleneck can be mitigated by computing distances over random sub-samples or mini-batches, retaining only the $k$ nearest neighbors per sample via approximate nearest-neighbor search, or adapting tree-based acceleration strategies such as Barnes-Hut~\citep{van2014accelerating} to the Poisson KL metric. As with \textit{t}-SNE and UMAP, the resulting embedding dimensions do not correspond to identifiable features or factors, which may limit interpretability of the embeddings, and the number of embedding dimensions $P$ must be specified by the user, although the optional $\ell_{1,2}$ penalty (App.~\ref{app:group_lasso}) offers a mechanism for automatic dimensionality selection. The similarity sharpness parameter $w$ and penalty weight $\gamma$ may subtly affect downstream performance; in our experiments, we checked varying degrees of $w$ and $\gamma$ (Figs.~\ref{fig:app_sweep_standard}-\ref{fig:app_sweep_xor}) and observed robust performance, with only a subtle effect on the recovered embeddings.

Future work includes extending \textit{p}-SNE to incorporate known label structure into the similarity computation, as in supervised or semi-supervised tasks, which could further improve embeddings when partial annotations are available. Another direction is to introduce a dynamics prior on latent state evolution when applied to temporal data.

\clearpage
\newpage

\acks{N.M. was funded as a fellow by the Kavli NeurData Discovery Institute. 
A.S.C. was supported in part by NSF CAREER award number 2340338.
Code and data are available at \url{https://github.com/NogaMudrik/PSNE-Poisson-Stochastic-Neighbor-Embedding}.
}

\bibliography{tmlr}
\newpage
\clearpage
\appendix
\setcounter{figure}{0}
\renewcommand{\thefigure}{S\arabic{figure}}
\setcounter{table}{0}
\renewcommand{\thetable}{S\arabic{table}}
\setcounter{equation}{0}
\renewcommand{\theequation}{S\arabic{equation}}
\appendix
{\Huge Appendix}

\section{Gradient Derivation}
\label{app:gradients}
We derive the gradient of $\mathcal{L}$ with respect to an embedding point $\bm{x}_n$. Let:
\begin{equation}
    F(\bm{X}) = \frac{1}{2}\sum_{n_1,n_2}\left(\sqrt{S_{n_1,n_2}} - \sqrt{Q_{n_1,n_2}}\right)^2
\end{equation}
so that $\mathcal{L} = \sqrt{F(\bm{X})}$. By the chain rule:
\begin{equation}
    \frac{\partial \mathcal{L}}{\partial \bm{x}_n} = \frac{1}{2\mathcal{L}}\frac{\partial F}{\partial \bm{x}_n}
\end{equation}
Differentiating $F$ with respect to $\bm{x}_n$, and noting that $S_{n_1,n_2}$ does not depend on $\bm{X}$:
\begin{equation}
    \frac{\partial F}{\partial \bm{x}_n} = -\frac{1}{2}\sum_{n_1,n_2} \frac{\sqrt{S_{n_1,n_2}} - \sqrt{Q_{n_1,n_2}}}{\sqrt{Q_{n_1,n_2}}} \frac{\partial Q_{n_1,n_2}}{\partial \bm{x}_n}
\end{equation}
\textbf{Gradient of $Q_{n_1,n_2}$.}
Let $u_{ab} = (1 + \|\bm{x}_a - \bm{x}_b\|^2)^{-1}$ and $Z = \sum_{a \neq b} u_{ab}$, so $Q_{ab} = u_{ab}/Z$. By the quotient rule:
\begin{equation}
    \frac{\partial Q_{ab}}{\partial \bm{x}_n} = \frac{1}{Z}\frac{\partial u_{ab}}{\partial \bm{x}_n} - Q_{ab}\frac{1}{Z}\frac{\partial Z}{\partial \bm{x}_n}
\end{equation}
The derivative of $u_{ab}$ with respect to $\bm{x}_n$ is nonzero only when $a = n$ or $b = n$:
\begin{equation}
    \frac{\partial u_{ab}}{\partial \bm{x}_n} = -2u_{ab}^2(\bm{x}_a - \bm{x}_b)(\delta_{an} - \delta_{bn})
\end{equation}
Since $u_{ab} = u_{ba}$, both cases contribute the same direction, giving:
\begin{equation}
    \frac{\partial Z}{\partial \bm{x}_n} = -4\sum_{j \neq n} u_{nj}^2 (\bm{x}_n - \bm{x}_j)
\end{equation}
Substituting and collecting terms, the gradient of $\mathcal{L}$ with respect to $\bm{x}_n$ is:
\begin{equation}
    \frac{\partial \mathcal{L}}{\partial \bm{x}_n} = \frac{1}{\mathcal{L}}\sum_{j \neq n}\left(A_{nj} - \beta\right)u_{nj}^2(\bm{x}_n - \bm{x}_j)
    \label{eq:gradient}
\end{equation}
where:
\begin{equation}
    A_{nj} = \frac{1}{2Z}\left(\frac{\sqrt{S_{nj}}-\sqrt{Q_{nj}}}{\sqrt{Q_{nj}}} + \frac{\sqrt{S_{jn}}-\sqrt{Q_{jn}}}{\sqrt{Q_{jn}}}\right)
\end{equation}
\begin{equation}
    \beta = \frac{1}{2Z}\sum_{n_1,n_2}\frac{\sqrt{S_{n_1,n_2}}-\sqrt{Q_{n_1,n_2}}}{\sqrt{Q_{n_1,n_2}}} Q_{n_1,n_2}
\end{equation}

\section{KL Divergence Between Poisson Distributions}
\label{app:kl_derivation}

The KL divergence between two distributions $\mathcal{U}$ and $\mathcal{V}$ is $\text{KL}(\mathcal{U}\|\mathcal{V}) = \sum_k \mathcal{U}(k) \log \frac{\mathcal{U}(k)}{\mathcal{V}(k)}$. For two Poisson distributions with means $\lambda_1$ and $\lambda_2$, $\mathcal{U}(k) = \frac{e^{-\lambda_1}\lambda_1^k}{k!}$ and $\mathcal{V}(k) = \frac{e^{-\lambda_2}\lambda_2^k}{k!}$, so:
\begin{equation}
    \text{KL}(\text{Pois}(\lambda_1)\|\text{Pois}(\lambda_2)) = \sum_{k=0}^{\infty} \mathcal{U}(k) \log \frac{e^{-\lambda_1}\lambda_1^k / k!}{e^{-\lambda_2}\lambda_2^k / k!} = \sum_{k=0}^{\infty} \mathcal{U}(k)\left[(\lambda_2 - \lambda_1) + k\log\frac{\lambda_1}{\lambda_2}\right]
\end{equation}
Applying $\sum_k \mathcal{U}(k) = 1$ and $\sum_k k\, \mathcal{U}(k) = \lambda_1$:
\begin{equation}
    = (\lambda_2 - \lambda_1) + \lambda_1\log\frac{\lambda_1}{\lambda_2} = \lambda_1\log\frac{\lambda_1}{\lambda_2} + \lambda_2 - \lambda_1
\end{equation}

\section{Robustness of the Hellinger Cost Under Sparse Similarities}
\label{app:robustness}
When the high-dimensional similarity distribution $\bm{S}$ is derived from sparse count data, many pairwise similarities $S_{n_1,n_2}$ are close to zero, and the corresponding low-dimensional similarities $Q_{n_1,n_2}$ may also approach zero during optimization. We show that the Hellinger cost remains well-behaved in this regime, whereas the KL divergence used in standard t-SNE does not.

Consider a pair $(n_1, n_2)$ with $S_{n_1,n_2} > 0$ and $Q_{n_1,n_2} \to 0$. Under the KL cost used in t-SNE, the contribution of this pair is:

\begin{equation}
S_{n_1,n_2} \log \frac{S_{n_1,n_2}}{Q_{n_1,n_2}} \to +\infty
\end{equation}

This divergence can dominate the total cost and produce large, unstable gradients, particularly when many pairs have near-zero $Q$ values.

Under the Hellinger cost (Eq.~\ref{eq:hellinger}), the contribution of the same pair is:

\begin{equation}
\left(\sqrt{S_{n_1,n_2}} - \sqrt{Q_{n_1,n_2}}\right)^2 \to S_{n_1,n_2}
\end{equation}

which is finite and bounded by $S_{n_1,n_2} \leq 1$. In the reverse case, where $S_{n_1,n_2} \to 0$ and $Q_{n_1,n_2} > 0$, the Hellinger contribution is $Q_{n_1,n_2}$, also finite. By contrast, the KL term $S_{n_1,n_2} \log \frac{S_{n_1,n_2}}{Q_{n_1,n_2}} \to 0$ in this direction, so KL is asymmetric in its sensitivity to the two types of mismatch.

In sparse count data, many sample pairs share few or no nonzero features, producing small $S_{n_1,n_2}$ values. During optimization, the corresponding $Q_{n_1,n_2}$ values may fluctuate near zero. The Hellinger cost handles these cases without numerical instability, whereas the KL cost requires clamping $Q$ away from zero, introducing an additional hyperparameter and potentially distorting gradients.

\section{Computational Complexity}
\label{app:complexity}

The dominant cost of \textit{p}-SNE is computing the pairwise Poisson KL distance matrix $\bm{D} \in \mathbb{R}^{N \times N}$, which requires $\mathcal{O}(N^2 M)$ operations for $N$ samples and $M$ features. Computing the joint probability matrix $\bm{S}$ from $\bm{D}$ requires $\mathcal{O}(N^2)$ operations (softmax normalization and symmetrization). Each gradient descent iteration involves updating the Student-t kernel matrix $\bm{Q}$ and computing the gradient (Eq.~\ref{eq:gradient}), both of which cost $\mathcal{O}(N^2 P)$ where $P$ is the embedding dimensionality. Over $K$ iterations, the total cost is $\mathcal{O}(N^2 M + K N^2 P)$. Since $M \gg P$ in all our experiments, the distance matrix computation dominates. For reference, on the largest dataset in this work (neural recordings, $N = 630$, $M = 200$, $P = 2$, $K = 500$), the full \textit{p}-SNE pipeline completes in under 30 seconds on a single CPU core.

\section{Group Lasso Extension}
\label{app:group_lasso}

\textit{p}-SNE supports an optional $\ell_{1,2}$ penalty on the embedding $\bm{X}$ that encourages entire embedding dimensions to be zeroed out, providing automatic selection of the effective embedding dimensionality. Let $\bm{x}^{(p)} \in \mathbb{R}^N$ denote the $p$-th row of $\bm{X}$, corresponding to the $p$-th embedding dimension across all samples. The augmented cost is:
\begin{equation}
    \mathcal{L}_\gamma(\bm{X}) = \mathcal{L} + \gamma \sum_{p=1}^{P} \|\bm{x}^{(p)}\|_2
    \label{eq:group_lasso}
\end{equation}
where $\gamma \geq 0$ controls the strength of the penalty. The penalty is a group lasso ($\ell_{1,2}$): it applies an $\ell_2$ norm within each embedding dimension (group) and an $\ell_1$ norm across dimensions, encouraging row-sparsity in $\bm{X}$. When $\|\bm{x}^{(p)}\|_2 \neq 0$, the gradient of the penalty with respect to $\bm{x}^{(p)}$ is:
\begin{equation}
    \frac{\partial}{\partial \bm{x}^{(p)}} \|\bm{x}^{(p)}\|_2 = \frac{\bm{x}^{(p)}}{\|\bm{x}^{(p)}\|_2}
\end{equation}
When $\|\bm{x}^{(p)}\|_2 = 0$, a subgradient of zero is used. The full gradient of $\mathcal{L}_\gamma$ follows by adding this term to Eq.~\ref{eq:gradient}. In practice, $\gamma$ is treated as a hyperparameter and tuned on a validation set or via grid search.

\begin{figure}
    \centering
    \includegraphics[width=0.8\linewidth]{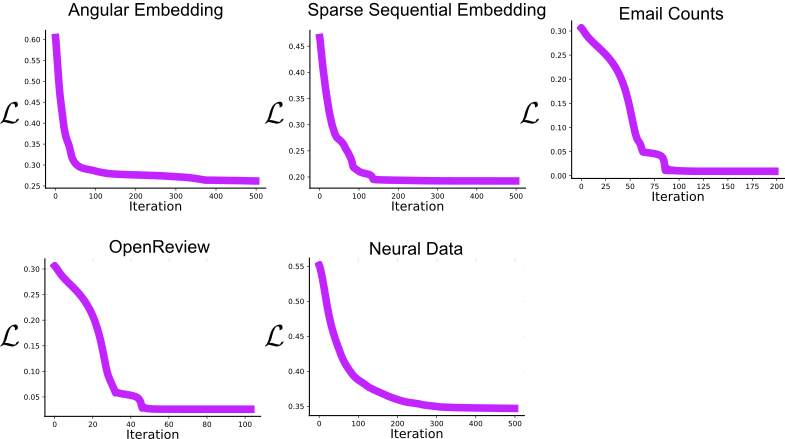}
    \caption{\textbf{\textit{p}-SNE cost convergence across datasets.}
\textbf{(A)} \textit{p}-SNE objective $\mathcal{L}$ as a function of gradient descent iteration for the two synthetic datasets (Angular Embedding and Sparse Sequential Embedding). $\mathcal{L}$ measures the weighted Hellinger distance between the high-dimensional joint probability matrix $\bm{S}$ and its low-dimensional approximation $\bm{Q}$; lower values indicate better structure preservation. Both datasets converge within 500 iterations.
\textbf{(B)} Cost convergence for real-world datasets (neural spike data and OpenReview).}
    \label{fig:cost}
\end{figure}

\begin{figure}
    \centering
    \includegraphics[width=1\linewidth]{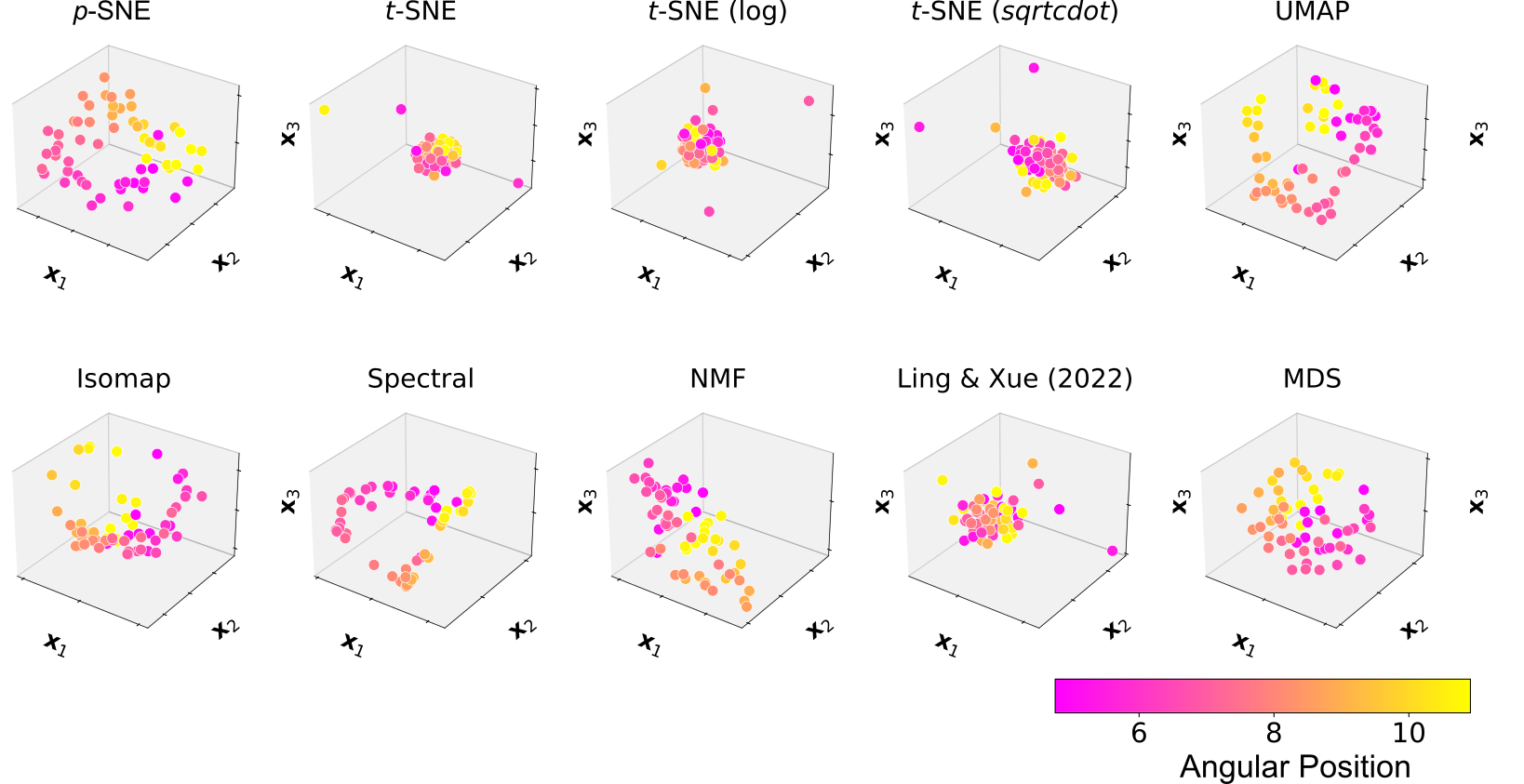}
    \caption{Angular Embedding: 3D embeddings colored by stimulus angular position. Same embeddings as Fig.~\ref{fig:synth_angular}D, now colored by the continuous angular coordinate of each stimulus on the underlying manifold. \textit{p}-SNE ($w$=1.0) recovers a smooth color gradient, indicating that it preserves the continuous manifold ordering beyond discrete group identity. Most baselines mix angular positions across the embedding.}
    \label{fig:synth_angular_by_t}
\end{figure}

\begin{figure}
    \centering
\includegraphics[width=1\linewidth]{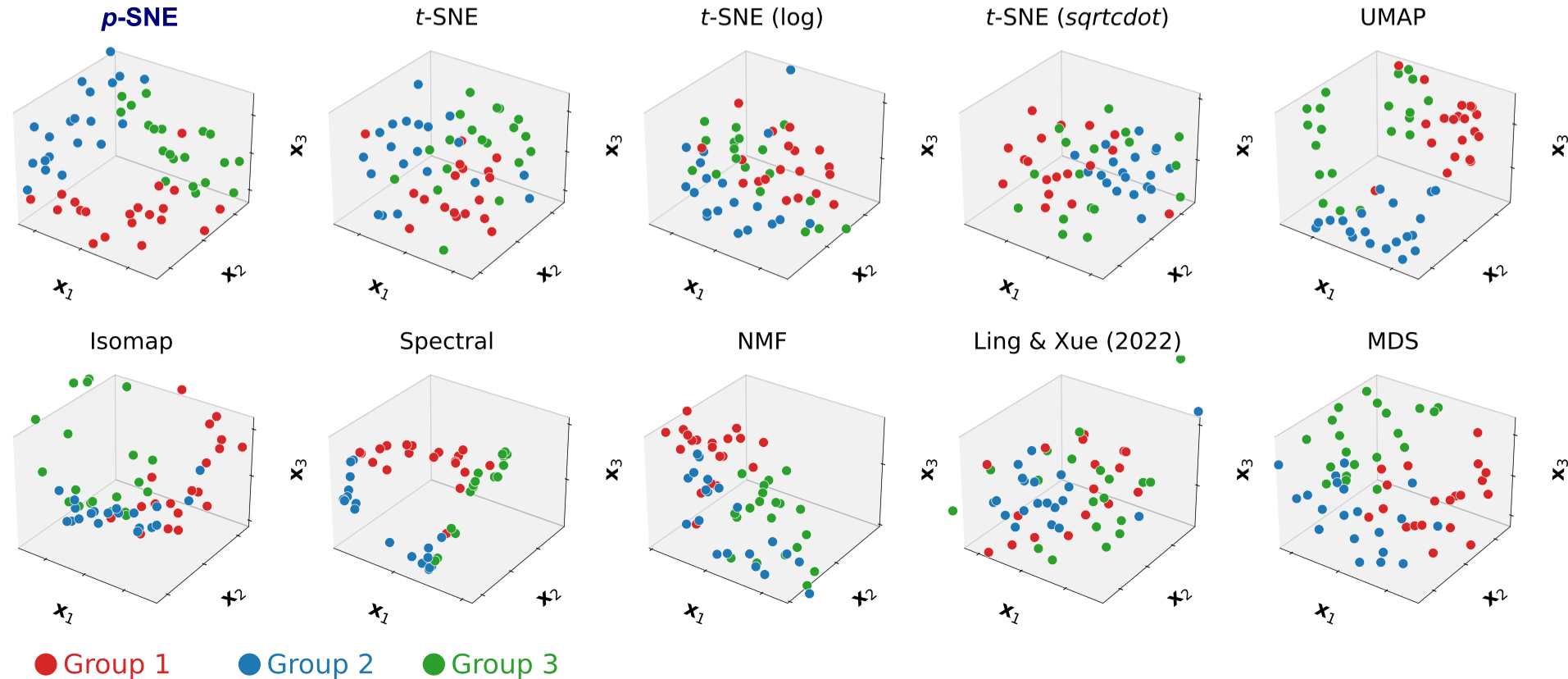}
    \caption{Angular Embedding: zoomed-in 3D embeddings. Same as Fig.~\ref{fig:synth_angular}D with axis limits set to the 10th-90th percentile range per method for improved visibility.}
    \label{fig:app_synth_standard_zoomed}
\end{figure}

\begin{figure}
    \centering
    \includegraphics[width=\linewidth]{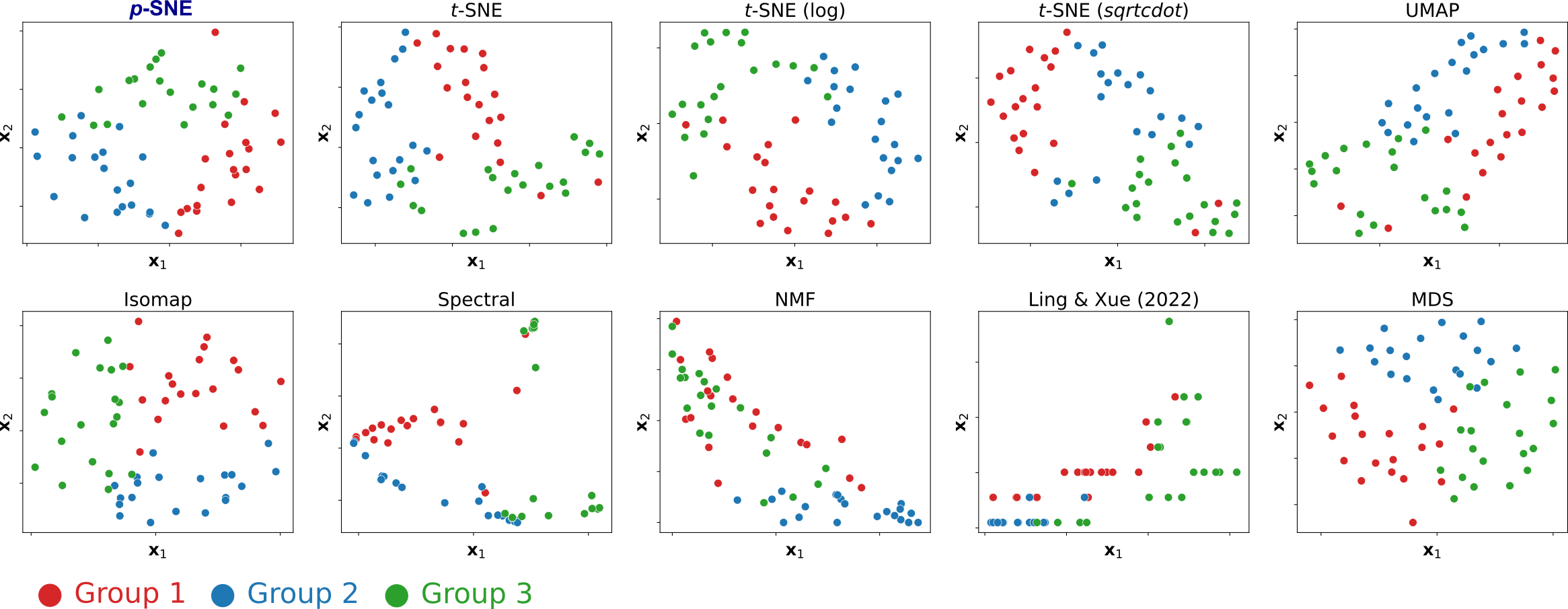}
    \caption{\textbf{Angular Embedding: 2D embeddings.} Two-dimensional embeddings of Poisson-distributed count data obtained by ten dimensionality reduction methods. Points are colored according to their position along the underlying manifold. \textit{p}-SNE ($w$=1.0) recovers a smooth, well-separated embedding that reflects the true manifold structure, while Isomap partially preserves the global ordering. The three $t$-SNE variants, UMAP, and NMF produce fragmented or mixed embeddings that fail to capture the global geometry. Spectral embedding, MDS, and the method of~\cite{ling2022dimension} achieve partial separation but exhibit distortion or overlap between neighboring regions of the manifold.}
    \label{fig:app_synth_standard_2d}
\end{figure}

\begin{figure}
    \centering
    \includegraphics[width=1\linewidth]{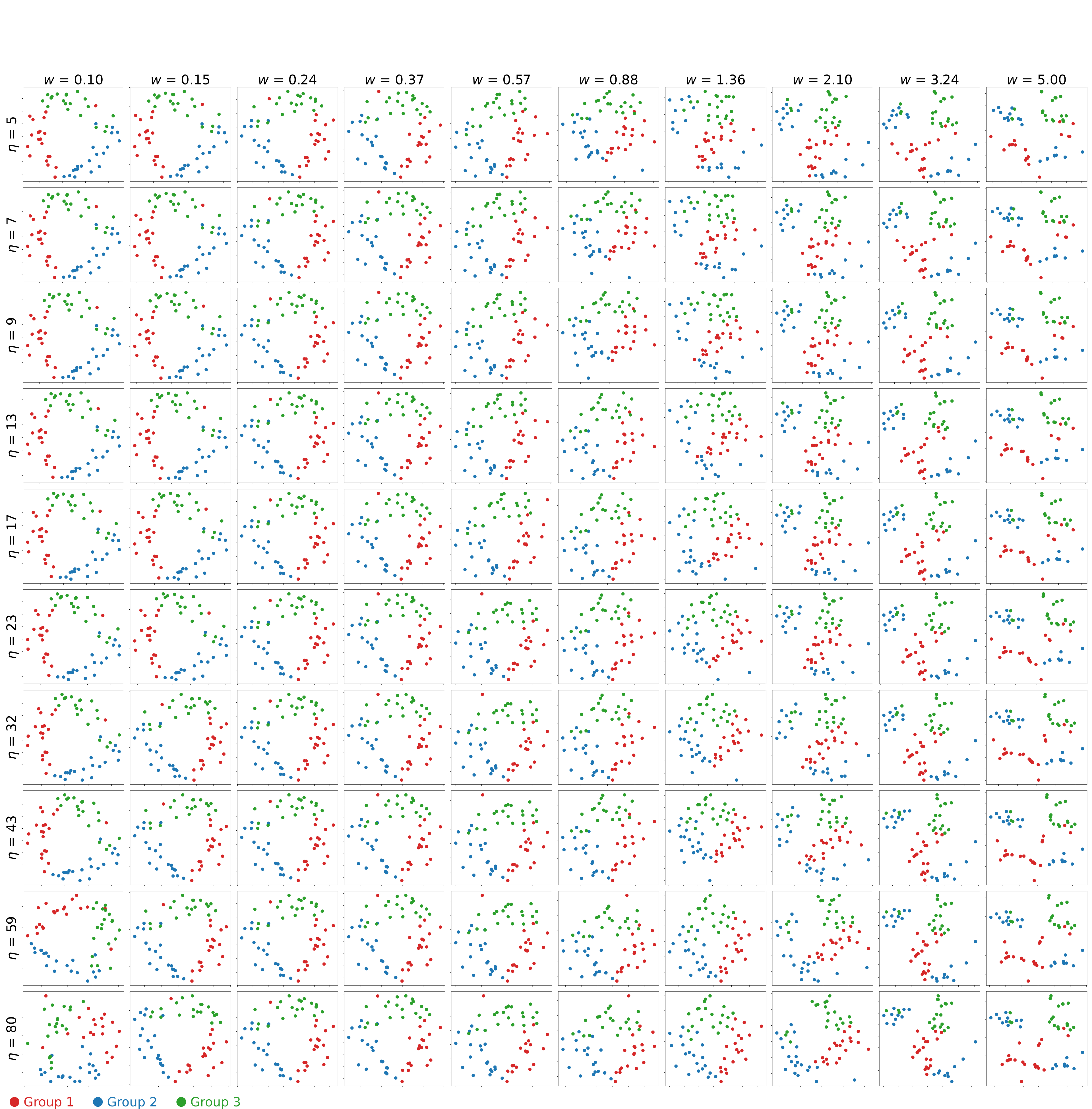}
    \caption{\textbf{Hyperparameter sensitivity: Angular Embedding.} 2D \textit{p}-SNE embeddings of the three-group Poisson count data across a grid of hyperparameter values: similarity sharpness $w$ (columns) and learning rate $\eta$ (rows). Points are colored by group identity. The embedding quality is stable across a wide range of $w$ and $\eta$ values, with group separation maintained for $w \in [0.10, 0.88]$ and $\eta \in [5, 80]$. At large $w$ ($\geq 1.36$), the similarity distribution becomes highly concentrated on nearest neighbors, causing the embedding to fragment. Increasing $\eta$ amplifies this effect. The results indicate that \textit{p}-SNE is robust to moderate variation in both hyperparameters.}
    \label{fig:app_sweep_standard}
\end{figure}

\begin{figure}
    \centering
    \includegraphics[width=1\linewidth]{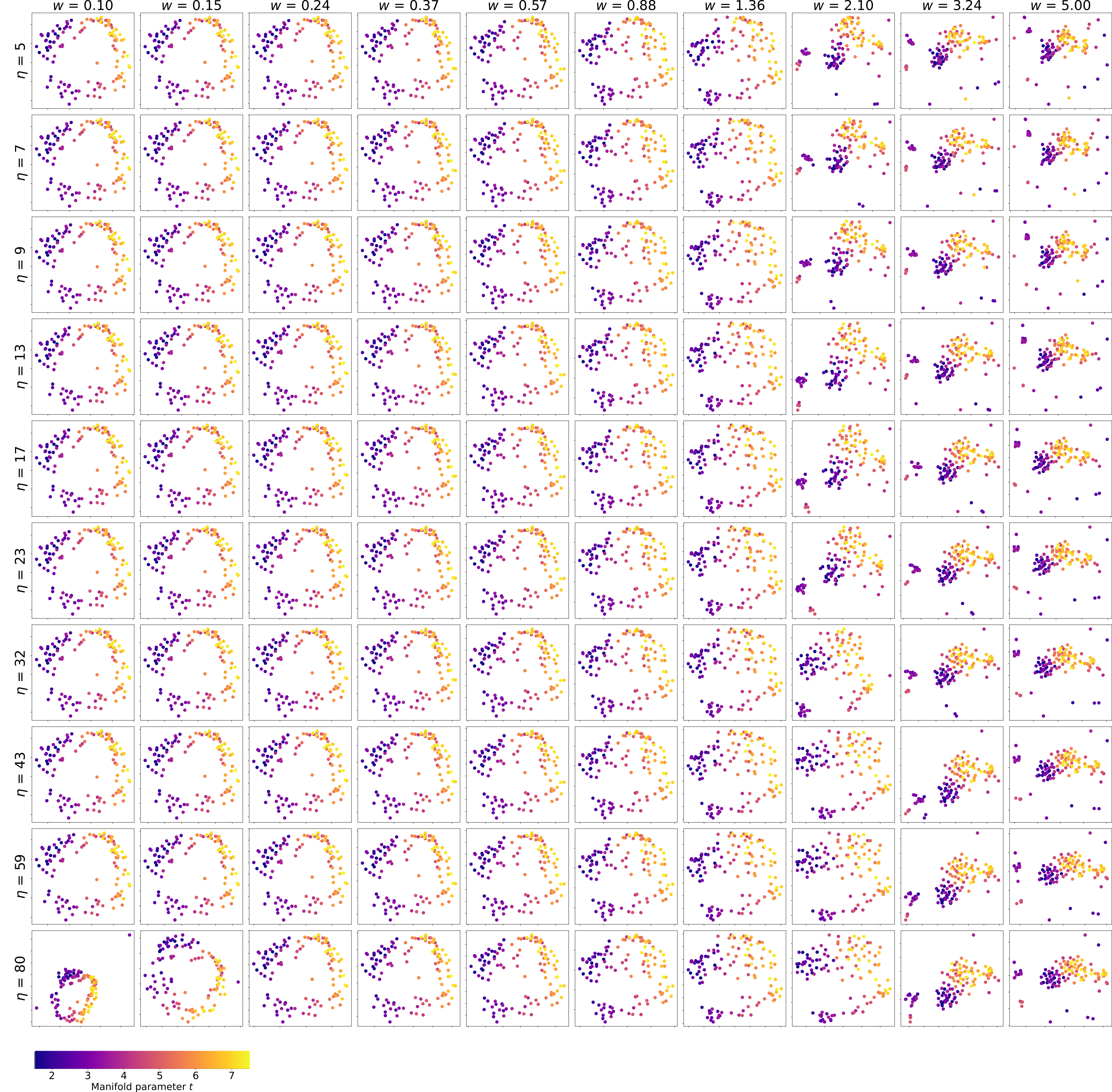}
    \caption{\textbf{Hyperparameter sensitivity: Sparse Sequential Embedding.} 2D \textit{p}-SNE embeddings of the four-segment Poisson manifold across the same $(w, \eta)$ grid as Fig.~\ref{fig:app_sweep_standard}. Points are colored by the continuous manifold parameter $t$. For low to moderate $w$ ($\leq 0.57$), the embedding recovers a smooth color gradient reflecting the underlying manifold structure across all tested $\eta$ values. At larger $w$ ($\geq 0.88$), the embedding progressively fragments into disconnected clusters, with adjacent manifold segments separating into distinct groups. This transition is consistent with the expected behavior: higher $w$ sharpens the similarity distribution, emphasizing local over global structure. The smooth-to-clustered transition is gradual, confirming that \textit{p}-SNE does not require precise hyperparameter tuning in the low-$w$ regime.}
    \label{fig:app_sweep_xor}
\end{figure}

\begin{figure}
    \centering
    \includegraphics[width=1\linewidth]{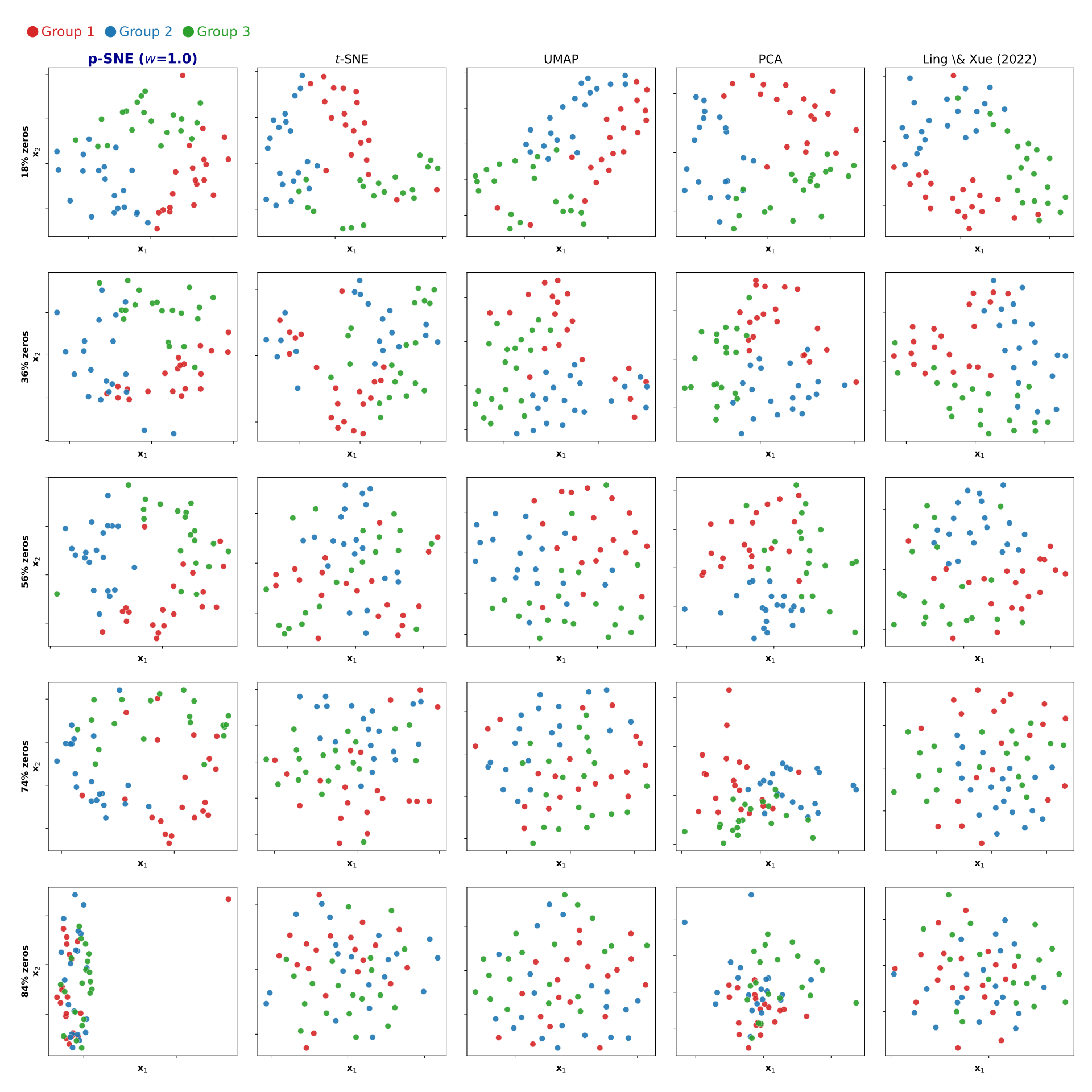}
    \caption{\textbf{Angular Embedding dataset: embeddings across sparsity levels.} Each row corresponds to a different sparsity level (18\%-84\% zeros), obtained by scaling the Poisson rate parameters. Columns show \textit{p}-SNE ($w{=}1.0$) and four representative baselines. At low sparsity all methods separate the three groups, but as sparsity increases, baselines progressively lose group structure while \textit{p}-SNE maintains clearer separation.}\label{fig:sparsity_1_app_baselines}
\end{figure}

\begin{figure}
    \centering
    \includegraphics[width=1\linewidth]{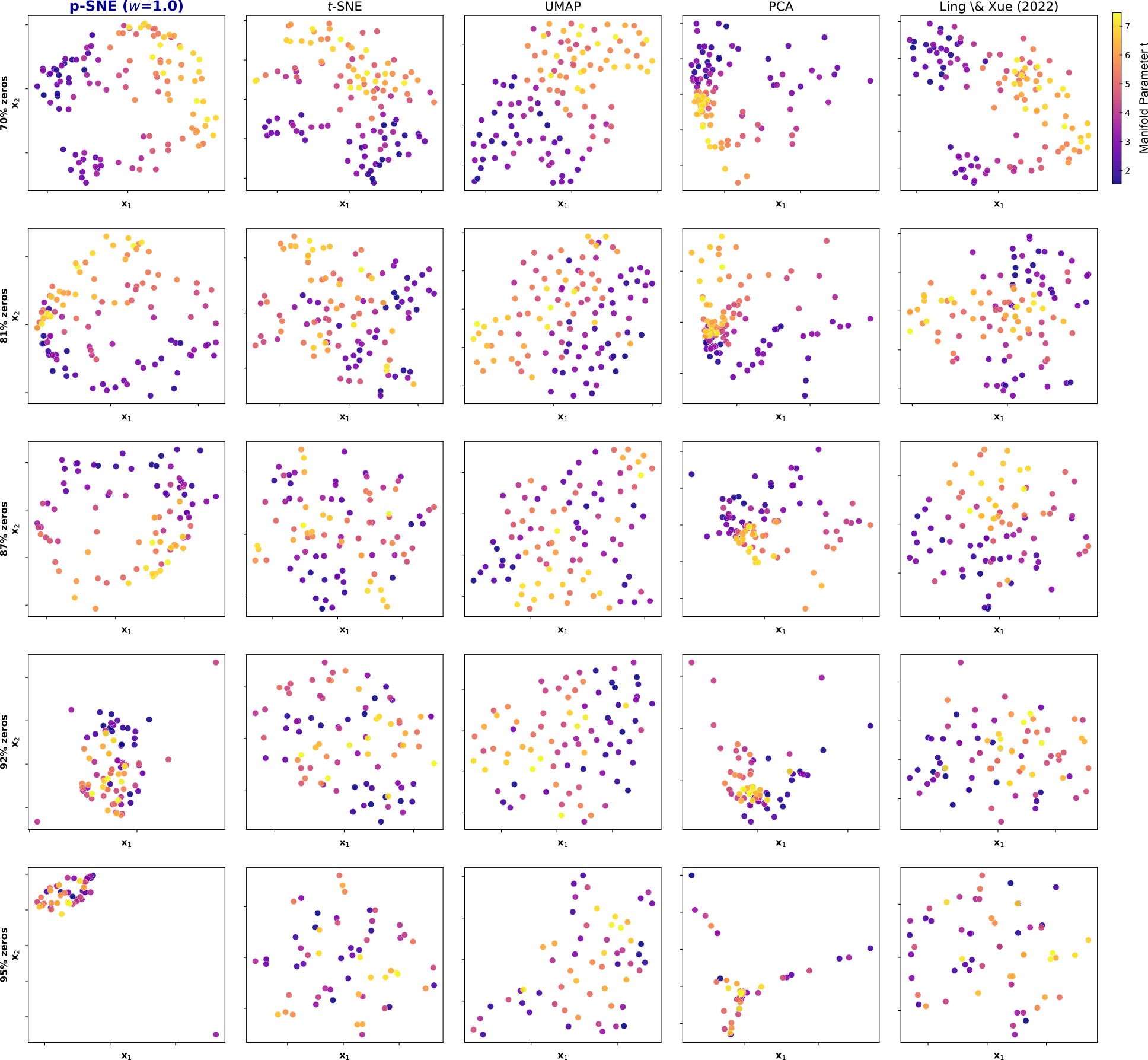}
   \caption{\textbf{Sparse Sequential Embedding dataset: embeddings across sparsity levels.} Each row corresponds to a different sparsity level (70\%-95\% zeros). Points are colored by the continuous manifold parameter $t$. Columns show \textit{p}-SNE ($w{=}1.0$) and four representative baselines. \textit{p}-SNE preserves the smooth color gradient at all sparsity levels, indicating recovery of the latent manifold ordering. Baselines, particularly PCA, lose this structure as sparsity increases.}
    \label{fig:sparsity_app_sequential}
\end{figure}

\begin{figure}
    \centering
    \includegraphics[width=0.5\linewidth]{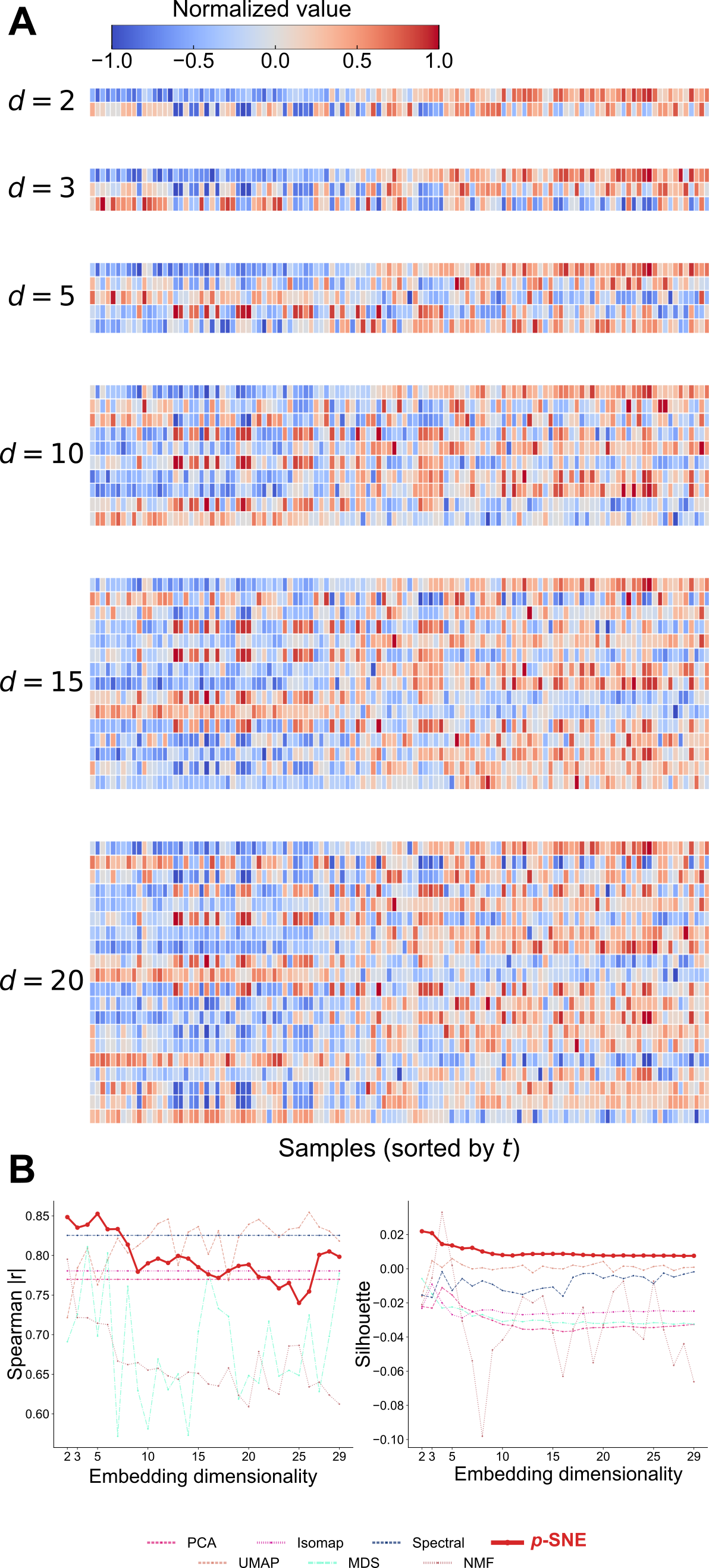}
    \caption{\textbf{Effect of embedding dimensionality on \textit{p}-SNE embeddings (Sparse Sequential Embedding).} \textbf{(A)}~\textit{p}-SNE embedding coordinates (normalized to $[-1,1]$) for $d \in \{2, 3, 5, 10, 15, 20\}$, with samples sorted by the latent manifold parameter $t$. Each row within a panel corresponds to one embedding dimension. At low $d$, \textit{p}-SNE captures the dominant smooth gradient along $t$; as $d$ increases, additional dimensions capture finer structure. \textbf{(B)}~Spearman $|r|$ (left) and silhouette score (right) as a function of embedding dimensionality for \textit{p}-SNE ($w{=}2.0$) and six baselines (PCA, UMAP, Isomap, Spectral, MDS, NMF). \textit{p}-SNE maintains the highest Spearman correlation across all dimensionalities, indicating consistent recovery of the latent manifold ordering. Data: 4-group Poisson count dataset ($n_{\mathrm{conds}}{=}30$, $n_{\mathrm{neurons}}{=}30$).}
    \label{fig:ndim_xor}
\end{figure}

\begin{figure}
    \centering
    \includegraphics[width=0.5\linewidth]{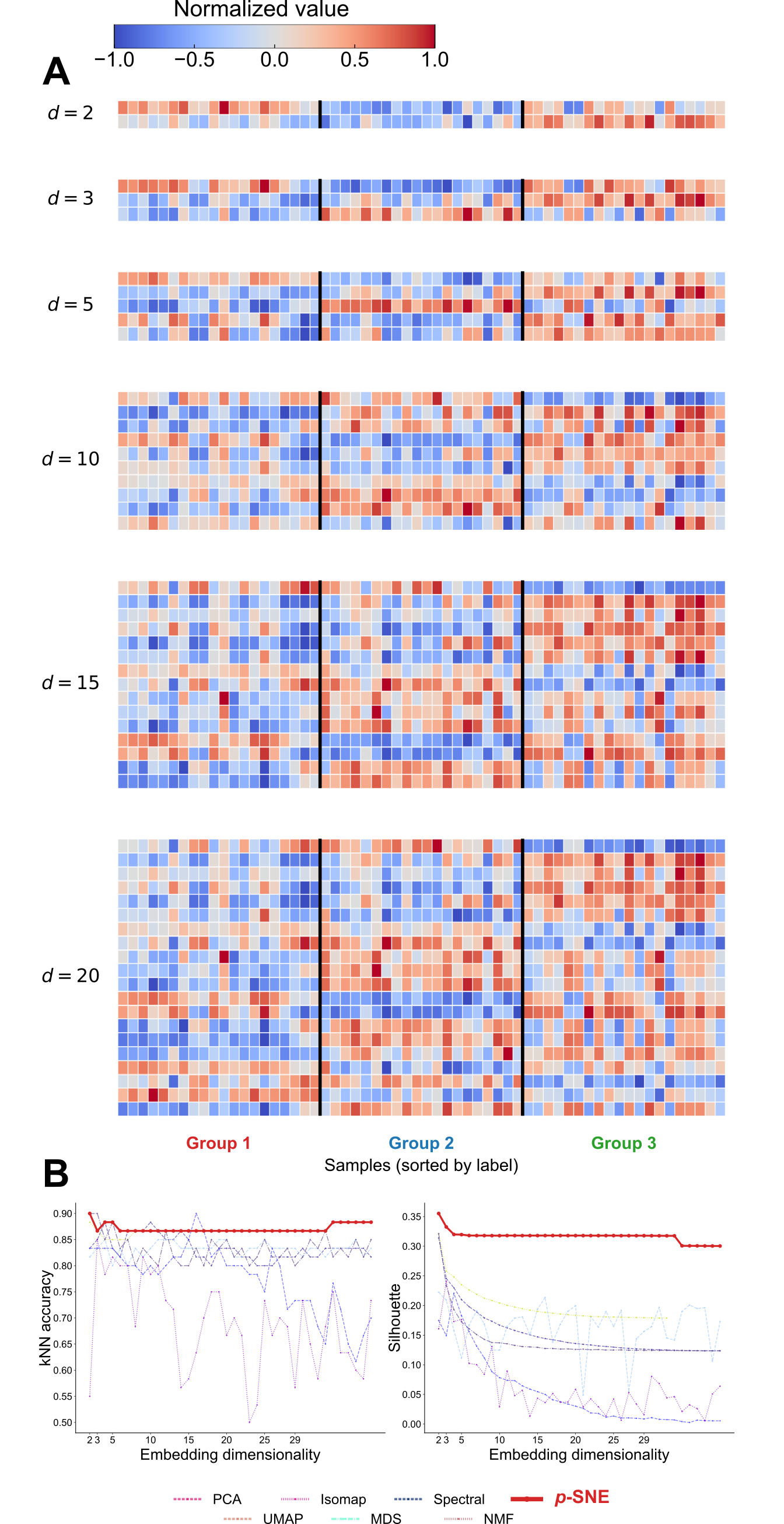}
    \caption{\textbf{Effect of embedding dimensionality on \textit{p}-SNE embeddings (Angular Embedding).} \textbf{(A)}~\textit{p}-SNE embedding coordinates (normalized to $[-1,1]$) for $d \in \{2, 3, 5, 10, 15, 20\}$, with samples sorted by group label. Dashed vertical lines mark group boundaries. At low $d$, \textit{p}-SNE already separates the three groups with distinct activation patterns per group; higher $d$ reveals additional within-group structure. \textbf{(B)}~kNN accuracy (left) and silhouette score (right) as a function of embedding dimensionality for \textit{p}-SNE ($w{=}2.0$) and six baselines. \textit{p}-SNE achieves the highest classification accuracy across dimensionalities. Data: 3-group Poisson count dataset ($n_{\mathrm{conds}}{=}20$, $n_{\mathrm{neurons}}{=}20$).}
    \label{fig:ndim_standard}
\end{figure}

\begin{figure}
    \centering
    \includegraphics[width=1\linewidth]{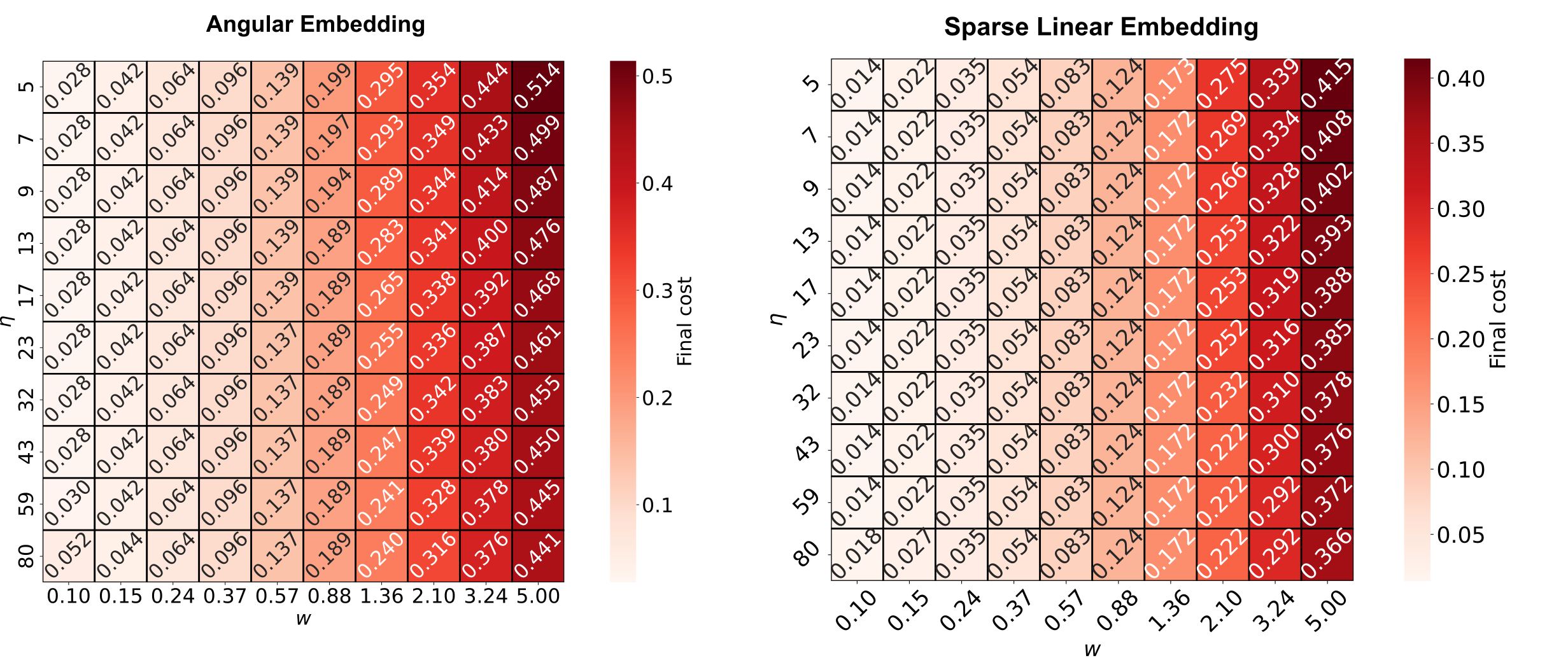}
    \caption{\textbf{Final \textit{p}-SNE cost across hyperparameters.} Hellinger cost $\mathcal{H}(S, Q)$ at convergence for the Angular Embedding (left) and Sparse Sequential Embedding (right) datasets, as a function of similarity sharpness $w$ (columns) and learning rate $\eta$ (rows). Lower cost (lighter) indicates better agreement between the high- and low-dimensional similarity distributions. For both datasets, cost increases monotonically with $w$, reflecting the increasingly concentrated similarity distribution. The cost is largely insensitive to $\eta$ for low to moderate $w$, confirming that the learning rate does not affect the quality of the converged solution in this regime.}
    \label{fig:app_cost_heatmap}
\end{figure}

\begin{figure}
    \centering
    \includegraphics[width=1\linewidth]{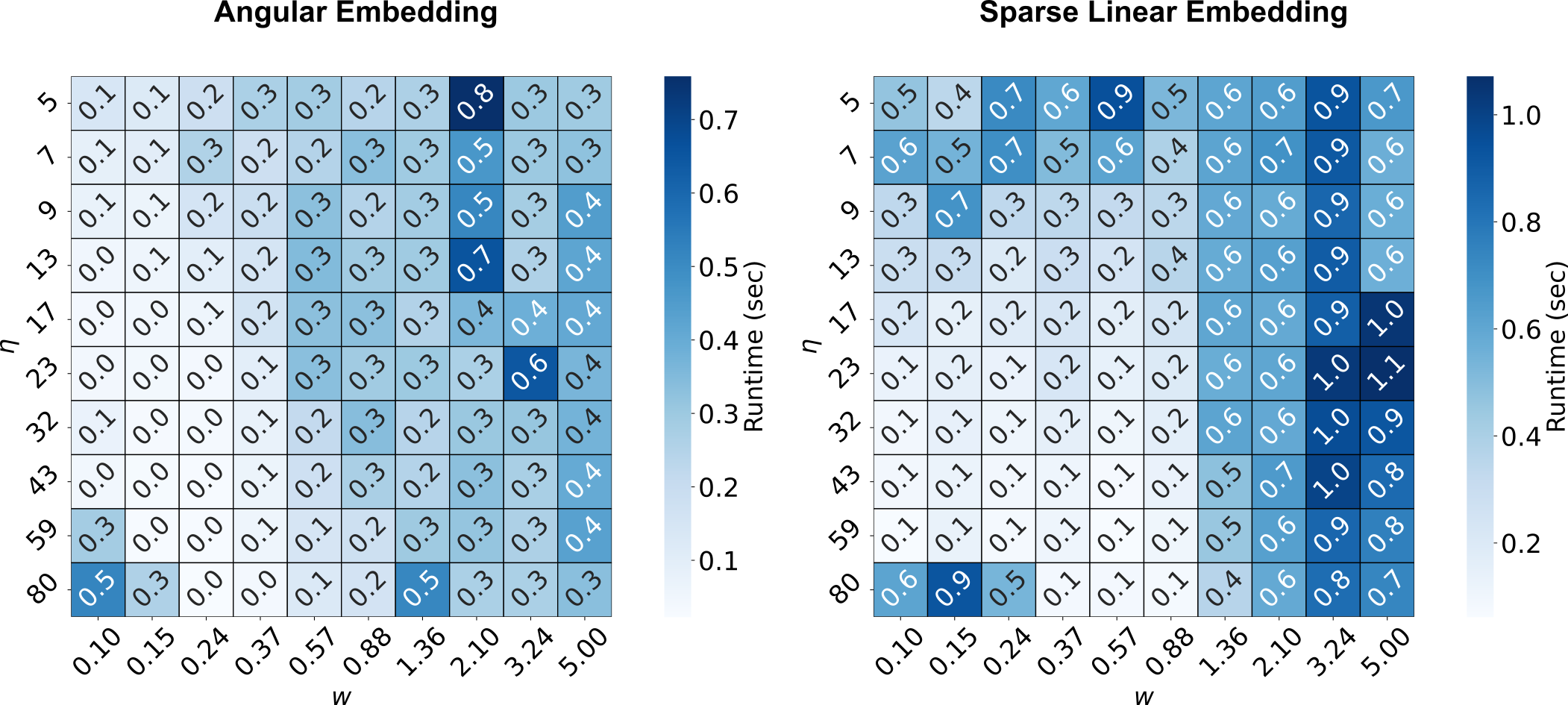}
    \caption{\textbf{\textit{p}-SNE runtime across hyperparameters.} Wall-clock runtime (seconds) for the same $(w, \eta)$ grid as Fig.~\ref{fig:app_cost_heatmap}. The Angular Embedding dataset ($60 \times 40$) runs in under 0.8 seconds for all configurations. The Sparse Sequential Embedding dataset ($119 \times 30$) takes up to 1.1 seconds at the most expensive settings. Runtime increases with both $w$ and $\eta$, as larger $w$ produces sharper gradients requiring more careful optimization steps, and larger $\eta$ produces larger updates that may slow convergence. Overall, \textit{p}-SNE runs in under one second on these small datasets across the full hyperparameter grid.}
    \label{fig:app_runtime_heatmap}
\end{figure}

\begin{figure}[t]
    \centering
    \includegraphics[width=1\linewidth]{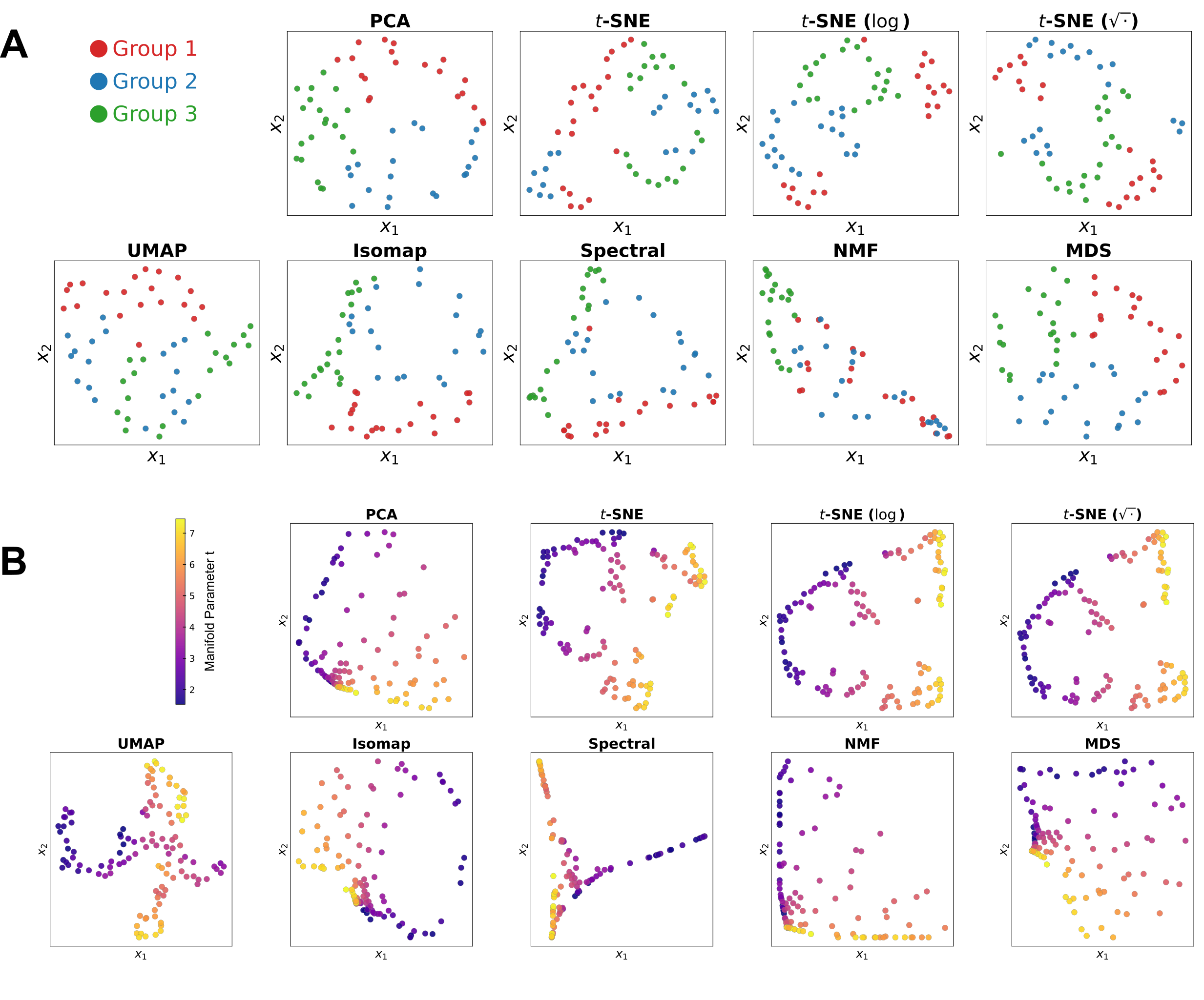}
    \caption{\textbf{Baseline embeddings applied to the noise-free rate matrix $\bm{\Lambda}$.} Nine baselines applied to the deterministic Poisson rate matrix $\bm{\Lambda}$ (before Poisson sampling) for both synthetic datasets. \textbf{(A)}~Angular Embedding ($\bm{\Lambda} \in \mathbb{R}^{40 \times 60}$, 3 groups), colored by group identity. \textbf{(B)}~Sparse Sequential Embedding ($\bm{\Lambda} \in \mathbb{R}^{30 \times 120}$, 4 groups), colored by the continuous manifold parameter $t$. When given access to the clean rates, several baselines recover structure similar to what \textit{p}-SNE achieves from the noisy Poisson counts (cf.\ Figs.~\ref{fig:synth_angular},~\ref{fig:synth_sparse_linear}), suggesting that \textit{p}-SNE effectively denoises the count observations and recovers the underlying rate geometry.}
    \label{fig:rate}
\end{figure}

\begin{figure}
    \centering
    \includegraphics[width=0.75\linewidth]{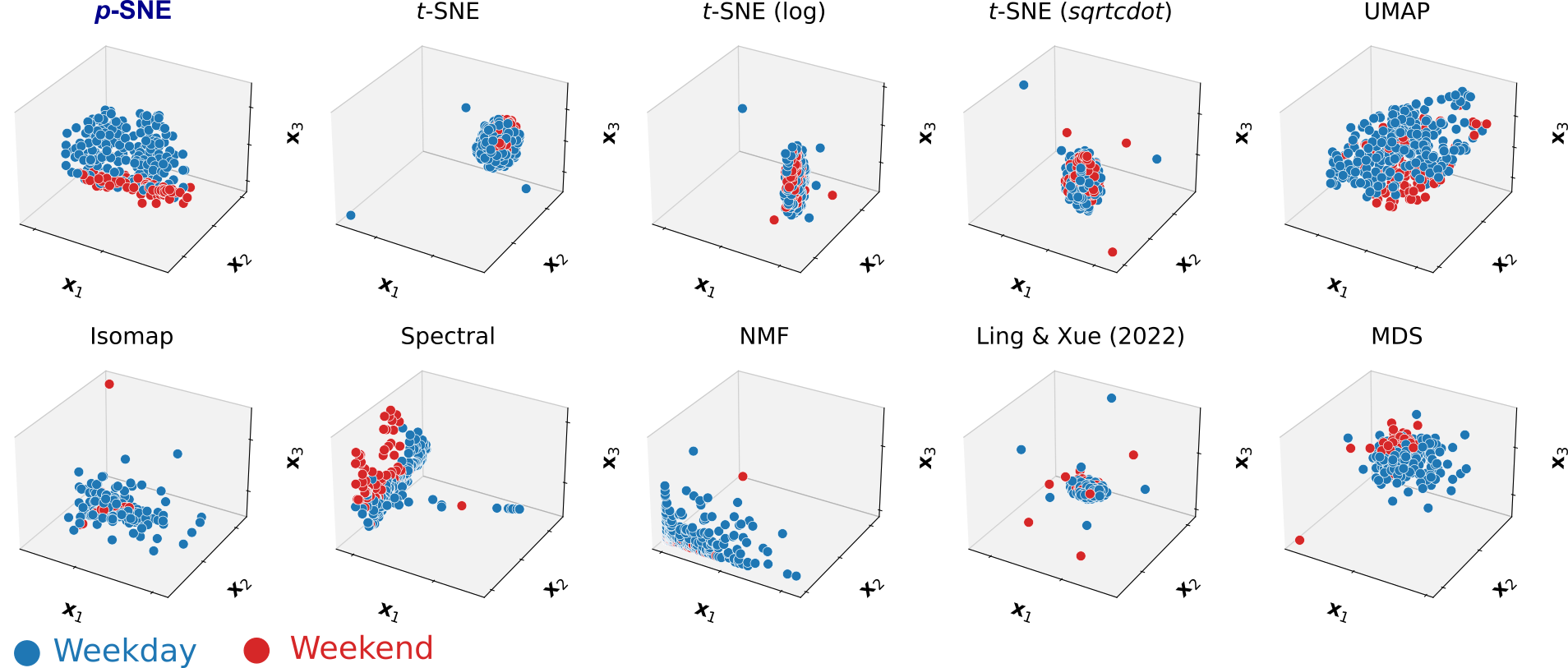}
    \caption{\textbf{Email data: binary weekend classification (all days).} 3D embeddings of the email count matrix colored by weekday (blue) vs.\ weekend (red), using all 370 days. \textit{p}-SNE ($w$=1.0) produces the clearest spatial separation between weekday and weekend email patterns, with weekend days consistently mapped to a distinct region of the embedding space. Most baselines show partial overlap, with $t$-SNE variants and UMAP mixing weekday and weekend days more extensively.}
    \label{fig:weekend_appendix_email}
\end{figure}

\section{About the Convergence Tolerance}\label{app:tau} 
The optimization terminates when the absolute change in cost between consecutive iterations falls below a tolerance $\tau > 0$. This parameter controls the trade-off between computational cost and solution precision. A smaller $\tau$ allows the optimizer to refine the embedding further, potentially improving fine-grained structure, but increases the number of iterations required. A larger $\tau$ stops optimization earlier, which may suffice when the cost curve has already plateaued but risks terminating before subtle structure is resolved. In practice, the cost curves across our experiments (Fig.\ref{fig:cost}) show that the Hellinger objective decreases rapidly during the first 100-200 iterations and plateaus well before the maximum iteration count. We use $\tau = 10^{-8}$. For larger datasets where runtime is a concern, $\tau = 10^{-5}$ provides a practical alternative with negligible effect on embedding quality.

\begin{table}[t]
\caption{\textbf{\textit{p}-SNE hyperparameters per experiment.} All experiments use momentum ($\mu_0=0.5$, $\mu_{\mathrm{final}}=0.8$, switch at iteration 250), early exaggeration (factor 4.0, 100 iterations), $\epsilon=10^{-2}$, $\gamma=0$, and random seed 42. For each experiment, we ran three variants ($w \in \{0.5, 1.0, 2.0\}$); the best variant column indicates the one shown in main text figures. The  \texorpdfstring{$w = 0.5$}{w=0.5} variant uses \texorpdfstring{$2\eta$}{2η}}.
\label{tab:psne_params}
\centering
\begin{small}
\begin{tabular}{lccccc}
\toprule
\textbf{Experiment} & $\bm{P}$ & $w$ (best) & $\eta$ & \textbf{max\_iter} & \textbf{Data size} ($N \times M$) \\
\midrule
Synthetic Angular       & 2 & 1.0 & 100 & 500 & $60 \times 40$ \\
Synthetic Sparse Sequential & 2 & 2.0 & 100 & 500 & $119 \times 30$ \\
Email                   & 3 & 1.0 & 100 & 500 & $153 \times 150$ \\
OpenReview              & 2 & 1.0 & 100 & 500 & $350 \times 100$ \\
Neural recordings       & 2 & 0.5 & 200 & 500 & $650 \times 200$ \\
\bottomrule
\end{tabular}
\end{small}
\end{table}

\begin{table}[ht]
\centering
\caption{Wall-clock runtime (seconds) for $d=2$ embedding dimensions on both synthetic datasets.}
\label{tab:runtime_d2}
\begin{tabular}{lcc}
\toprule
Method & Angular Embedding & Sparse Sequential \\
\midrule
PCA & 0.27 & 0.07 \\
NMF & 0.40 & 0.06 \\
Isomap & 0.51 & 0.10 \\
Spectral & 0.67 & 0.06 \\
\textit{p}-SNE ($w$=0.5) & 1.07 & 1.26 \\
Ling \& Xue (2022) & 2.12 & 0.55 \\
\textit{p}-SNE ($w$=1.0) & 2.23 & 0.32 \\
\textit{p}-SNE ($w$=2.0) & 4.36 & 0.71 \\
$t$-SNE ($\sqrt{\cdot}$) & 4.90 & 0.53 \\
$t$-SNE & 6.30 & 0.55 \\
$t$-SNE ($\log$) & 6.43 & 0.67 \\
MDS & 9.04 & 1.07 \\
UMAP & 24.08 & 2.90 \\
\bottomrule
\end{tabular}
\end{table}

\begin{table}[ht]
\centering
\caption{Wall-clock runtime (seconds) for $d=3$ embedding dimensions on both synthetic datasets.}
\label{tab:runtime_d3}
\begin{tabular}{lcc}
\toprule
Method & Angular Embedding & Sparse Sequential \\
\midrule
PCA & 0.18 & 0.04 \\
NMF & 0.21 & 0.07 \\
Spectral & 0.27 & 0.09 \\
Isomap & 0.59 & 0.07 \\
$t$-SNE ($\log$) & 0.99 & 0.84 \\
Ling \& Xue (2022) & 1.65 & 1.45 \\
\textit{p}-SNE ($w$=0.5) & 2.16 & 0.85 \\
MDS & 2.17 & 1.37 \\
\textit{p}-SNE ($w$=1.0) & 2.43 & 0.59 \\
$t$-SNE ($\sqrt{\cdot}$) & 2.54 & 0.85 \\
\textit{p}-SNE ($w$=2.0) & 3.15 & 0.96 \\
$t$-SNE & 7.69 & 0.79 \\
UMAP & 24.40 & 2.51 \\
\bottomrule
\end{tabular}
\end{table}

\begin{table}[h]
\centering
\caption{Wall-clock runtime (seconds) across embedding dimensions $d \in \{2, 5, 10, 15, 20\}$ on the Angular Embedding dataset ($N=60$, $M=40$).}
\label{tab:runtime_dims_angular}
\begin{tabular}{lccccc}
\toprule
Method & $d$=2 & $d$=5 & $d$=10 & $d$=15 & $d$=20 \\
\midrule
PCA & 0.27 & 0.06 & 0.04 & 0.10 & 0.08 \\
NMF & 0.40 & 0.05 & 0.07 & 0.14 & 0.15 \\
Isomap & 0.51 & 0.08 & 0.06 & 0.07 & 0.05 \\
Spectral & 0.67 & 0.09 & 0.04 & 0.07 & 0.06 \\
\textit{p}-SNE ($w$=0.5) & 1.07 & 1.31 & 0.66 & 1.21 & 3.25 \\
Ling \& Xue (2022) & 2.12 & 1.08 & 0.64 & 1.18 & 0.98 \\
\textit{p}-SNE ($w$=1.0) & 2.23 & 1.89 & 0.55 & 1.21 & 1.13 \\
\textit{p}-SNE ($w$=2.0) & 4.36 & 1.19 & 0.66 & 2.30 & 1.03 \\
$t$-SNE ($\sqrt{\cdot}$) & 4.90 & 1.29 & 0.65 & 0.85 & 0.96 \\
$t$-SNE & 6.30 & 0.87 & 0.87 & 1.27 & 0.93 \\
$t$-SNE ($\log$) & 6.43 & 1.00 & 0.60 & 1.13 & 0.97 \\
MDS & 9.04 & 1.26 & 0.80 & 1.23 & 1.00 \\
UMAP & 24.08 & 4.86 & 2.24 & 4.11 & 3.21 \\
\bottomrule
\end{tabular}
\end{table}

\begin{table}[h]
\centering
\caption{Wall-clock runtime (seconds) across embedding dimensions $d \in \{2, 5, 10, 15, 20\}$ on the Sparse Sequential Embedding dataset ($N=119$, $M=30$).}
\label{tab:runtime_dims_xor}
\begin{tabular}{lccccc}
\toprule
Method & $d$=2 & $d$=5 & $d$=10 & $d$=15 & $d$=20 \\
\midrule
NMF & 0.06 & 0.07 & 0.09 & 0.12 & 0.09 \\
Spectral & 0.06 & 0.07 & 0.07 & 0.11 & 0.08 \\
PCA & 0.07 & 0.06 & 0.05 & 0.07 & 0.09 \\
Isomap & 0.10 & 0.10 & 0.11 & 0.10 & 0.09 \\
\textit{p}-SNE ($w$=1.0) & 0.32 & 0.77 & 2.34 & 3.33 & 4.54 \\
$t$-SNE ($\sqrt{\cdot}$) & 0.53 & 1.46 & 1.52 & 1.65 & 1.54 \\
$t$-SNE & 0.55 & 1.51 & 1.35 & 1.99 & 1.85 \\
Ling \& Xue (2022) & 0.55 & 1.50 & 1.50 & 1.60 & 1.65 \\
$t$-SNE ($\log$) & 0.67 & 1.66 & 1.81 & 1.71 & 1.72 \\
\textit{p}-SNE ($w$=2.0) & 0.71 & 1.34 & 2.29 & 3.83 & 5.02 \\
MDS & 1.07 & 1.35 & 1.27 & 1.40 & 1.42 \\
\textit{p}-SNE ($w$=0.5) & 1.26 & 1.03 & 2.45 & 4.29 & 4.00 \\
UMAP & 2.90 & 2.78 & 2.51 & 3.33 & 2.87 \\
\bottomrule
\end{tabular}
\end{table}

\section{Data Generation and Data Preprocessing Details}
\label{app:preprocessing}

\subsection{Synthetic Data}

\subsubsection{Angular Embedding}\label{app:data_generation_synth1}

The Angular Embedding dataset simulates a neural population recording on a nonlinear manifold. The generative process proceeds as follows.

\paragraph{Stimulus positions.} The manifold is parameterized by two coordinates: an angular parameter $t$ and a height $h$. We define three stimulus classes $g \in \{0, 1, 2\}$, each occupying a contiguous interval along $t$:

\begin{equation}
    t \sim \mathrm{Uniform}\!\left(\frac{2\pi g}{3} + \frac{3\pi}{2},\; \frac{2\pi(g+1)}{3} + \frac{3\pi}{2}\right), \quad h \sim \mathrm{Uniform}(0, 10)
\end{equation}
For each class, we draw 20 stimulus positions $(t_i, h_i)$, yielding $N = 60$ trials total.

\paragraph{Neuron preferred positions.} Each neuron $m \in \{0, \ldots, M{-}1\}$ with $M = 40$ has a preferred position $(t_m^*, h_m^*)$ on the manifold, defined as:
\begin{equation}
    t_m^* = \frac{3\pi}{2} + \frac{2\pi\,(m \bmod 20)}{20}, \quad h_m^* = \frac{10 \cdot \lfloor m / 20 \rfloor}{2}
\end{equation}
Neurons with adjacent indices thus have nearby preferred positions along $t$.

\paragraph{Firing rates and spike counts.} The firing rate of neuron $m$ on trial $i$ depends on the distance between the stimulus position $(t_i, h_i)$ and the neuron's preferred position $(t_m^*, h_m^*)$:
\begin{equation}
    \lambda_{i,m} = \lambda_{\mathrm{bias}} + \lambda_{\mathrm{peak}} \exp\!\left(-\frac{\Delta t_{i,m}^2}{2} - \frac{\Delta h_{i,m}^2}{20}\right)
\end{equation}
where $\Delta h_{i,m} = |h_i - h_m^*|$ and $\Delta t_{i,m} = \min(|t_i - t_m^*|,\; 2\pi - |t_i - t_m^*|)$ accounts for the circular topology of the angular coordinate. We set $\lambda_{\mathrm{bias}} = 1.0$ and $\lambda_{\mathrm{peak}} = 8.0$, so rates range from 1.0 to 9.0. The observed spike count is drawn as $Y_{i,m} \sim \mathrm{Poisson}(\lambda_{i,m})$. Random seed 42 is used throughout.

\subsubsection{Sparse Sequential Embedding}\label{app:data_generation_synth2}

The Sparse Sequential Embedding dataset follows a similar generative process but with lower firing rates to produce high sparsity.

\paragraph{Stimulus positions.} We define four stimulus classes $g \in \{0, 1, 2, 3\}$, each occupying a contiguous interval along a linear parameter $t$:
\begin{equation}
    t \sim \mathrm{Uniform}\!\left(1.5g + 1.5,\; 1.5(g+1) + 1.5\right), \quad h \sim \mathrm{Uniform}(0, 5)
\end{equation}
For each class, we draw 30 stimulus positions $(t_i, h_i)$, yielding $N = 120$ trials (one all-zero sample was removed, giving 119).

\paragraph{Neuron preferred positions.} Each neuron $m \in \{0, \ldots, M{-}1\}$ with $M = 30$ has a preferred position $(t_m^*, h_m^*)$:
\begin{equation}
    t_m^* = 1.5 + \frac{6.0\,(m \bmod 25)}{25}, \quad h_m^* = \frac{5.0 \cdot \lfloor m / 25 \rfloor}{2}
\end{equation}

\paragraph{Firing rates and spike counts.} The firing rate of neuron $m$ on trial $i$ is:
\begin{equation}
    \lambda_{i,m} = \lambda_{\mathrm{bias}} + \lambda_{\mathrm{peak}} \exp\!\left(-\frac{d_{i,m}^2}{3}\right)
\end{equation}
where $d_{i,m} = \sqrt{(t_i - t_m^*)^2 + (h_i - h_m^*)^2}$ is the Euclidean distance between the stimulus and the neuron's preferred position. We set $\lambda_{\mathrm{bias}} = 0.1$ and $\lambda_{\mathrm{peak}} = 2.5$, so rates range from 0.1 to 2.6. The observed spike count is drawn as $Y_{i,m} \sim \mathrm{Poisson}(\lambda_{i,m})$. Random seed 42 is used throughout.

\subsection{Email Communication Data}\label{app:email_preproc}
We extracted email metadata from a personal Microsoft Outlook archive spanning October 2024 to February 2026 (428 days at daily resolution). Preprocessing proceeded as follows: (1)~we removed emails from the account holder to exclude self-sent messages; (2)~we removed emails from specific non-personal senders (e.g., newspaper newsletters); (3)~we selected the top 150 most frequent senders by total email count; (4)~sender identities were anonymized by replacing real names with pseudonyms (e.g., ``person\_A'', ``university\_A\_newsletter''); (5)~we randomly subsampled 370 out of 428 available days (seed 42), preserving the natural weekday/weekend ratio. The resulting count matrix $\bm{Y} \in \mathbb{Z}_{\geq 0}^{370 \times 150}$ records how many emails each sender sent on each day. For the three-class evaluation (Fig.~\ref{fig:email}), we retained only Tuesdays, Saturdays, and Sundays (153 days).

\subsection{OpenReview Academic Paper Abstracts}\label{app:openreview}
We scraped paper metadata and abstracts from ICLR~2024, ICLR~2025, and TMLR via the OpenReview API. We retained only papers with resolved decisions (accepted, rejected, or withdrawn), excluding papers still under review. 
We assigned each paper to one of twelve research areas via keyword matching on the lowercased abstract text. Table~\ref{tab:area_keywords} lists the keywords used for each area. Papers matching zero or multiple areas were excluded. For the main experiment, we retained five areas (Diffusion, GNN, LLM, Neuroscience, and RL) with 70 papers each, yielding 350 papers total.
\begin{table}[h!]
\centering
\caption{Research area keyword definitions for the OpenReview experiment.}
\label{tab:area_keywords}
\small
\begin{tabular}{ll}
\toprule
\textbf{Area} & \textbf{Keywords} \\
\midrule
Diffusion & diffusion model, score-based, score matching \\
GNN & gnn, node classification, link prediction, subgraph \\
LLM & llm, gpt-4, gpt-3, chatgpt, gemini, claude, instruction tuning, in-context learning, chain of thought \\
Neuroscience & brain, neuroscience, fmri, eeg, cortex, neuroimaging \\
RL & reinforcement learning, policy gradient, multi-agent, offline rl, online rl, atari, mujoco \\
\bottomrule
\end{tabular}
\end{table}
We applied \texttt{sklearn.feature\_extraction.text.CountVectorizer} with English stopwords, WordNet lemmatization, a minimum document frequency of 5, and a maximum document frequency of 50\%. From the resulting vocabulary, we selected the top 500 words by total count. We then selected the 100 most frequent words after removing words appearing in more than 50\% of documents (e.g., data, method, model, etc.), yielding $\bm{Y} \in \mathbb{Z}_{\geq 0}^{350\times 100}$.

\subsection{Neural Recordings Data}
We used a single session (sub-719828686, ses-754312389) from the Allen Brain Observatory Visual Coding Neuropixels dataset (accessed via the DANDI archive, dandiset 000021). Spike times were binned into 50\,ms windows within each of the 2-second trial. We retained only units classified as ``good'' quality. For the trials-summed representation, we summed spike counts over all 40 time bins per trial, yielding one count vector per trial. We then removed near-silent neurons with fewer than 2 total spikes across all trials, then selected the 200 least active neurons. We removed trials with all-zero spike vectors, such that the final matrix is $\bm{Y} \in \mathbb{Z}_{\geq 0}^{630 \times 200}$.

\end{document}